\documentclass{article}

     \usepackage[preprint]{neurips_2021}

\usepackage[utf8]{inputenc} % allow utf-8 input
\usepackage[T1]{fontenc}    % use 8-bit T1 fonts
\usepackage{hyperref}       % hyperlinks
\usepackage{url}            % simple URL typesetting
\usepackage{booktabs}       % professional-quality tables
\usepackage{amsfonts}       % blackboard math symbols
\usepackage{nicefrac}       % compact symbols for 1/2, etc.
\usepackage{microtype}      % microtypography
\usepackage{xcolor}         % colors
\usepackage{amssymb}
\usepackage{subfigure}
\usepackage{amsmath}

\usepackage{graphicx}       % used by scalebox (among others)

\title{Drawing Multiple Augmentation Samples Per Image During Training Efficiently Decreases Test Error}
\author{%
  Stanislav Fort\thanks{Work performed while interning at DeepMind.} \\
  Stanford University\\
  \texttt{stanislav.fort@gmail.com} \\
  \And
  Andrew Brock  \\
  DeepMind \\
   \texttt{ajbrock@deepmind.com} \\
   \AND
  Razvan Pascanu \\
  DeepMind \\
   \texttt{razp@deepmind.com} \\
   \And
  Soham De \\
  DeepMind \\
   \texttt{sohamde@deepmind.com} \\
   \And
   Samuel L. Smith\thanks{Corresponding author} \\
  DeepMind \\
   \texttt{slsmith@deepmind.com} \\
}

\begin{document}

\maketitle

\begin{abstract}
In computer vision, it is standard practice to draw a single sample from the data augmentation procedure for each unique image in the mini-batch. However recent work has suggested drawing multiple samples can achieve higher test accuracies. In this work, we provide a detailed empirical evaluation of how the number of augmentation samples per unique image influences model performance on held out data when training deep ResNets. We demonstrate drawing multiple samples per image consistently enhances the test accuracy achieved for both small and large batch training. Crucially, this benefit arises even if different numbers of augmentations per image perform the same number of parameter updates and gradient evaluations (requiring the same total compute). Although prior work has found variance in the gradient estimate arising from subsampling the dataset has an implicit regularization benefit, our experiments suggest variance which arises from the data augmentation process harms generalization. We apply these insights to the highly performant NFNet-F5, achieving 86.8$\%$ top-1 w/o extra data on ImageNet.
\end{abstract}

\section{Introduction}

Data augmentation plays a crucial role in computer vision, and it is currently essential to achieve competitive performance on held-out data \citep{shorten2019survey}. However the origin of the benefits of data augmentation are not fully understood. In addition, a number of authors have identified that SGD has an implicit regularization benefit, whereby large learning rates and small batch sizes achieve higher accuracy on held-out data \citep{keskar2016large,mandt2017stochastic, smith2017bayesian, jastrzkebski2017three, chaudhari2018stochastic, li2019towards, smith2021origin}, yet most authors have not considered how data augmentation and large learning rates interact during training. 

In this work, we note that data augmentation has two distinct influences on the gradient of a single training example. First, augmentation operations like left-right flips, random crops or RandAugment \citep{cubuk2020randaugment} change the expected value of the gradient of an example, introducing bias. Second, since we draw a finite number of samples from the augmentation procedure in each minibatch (typically one per unique image), data augmentation also introduces variance. A key goal of this work is  to establish whether the benefits of data augmentation arise solely from bias, or whether the variance introduced by the augmentation procedure is also beneficial. We were inspired to study this question by a recent study of Dropout \citep{wei2020implicit}, which claimed that both the bias and variance introduced by Dropout contribute to generalization. In addition, we wish to understand how the variance introduced by the augmentation procedure interacts with the variance introduced when we estimate the gradient on a subset of the training set.

To distinguish between the roles of bias and variance, we exploit a simple modification to standard training pipelines, which we call augmentation multiplicity (see Figure \ref{fig:sketch}). This technique was first proposed by \citet{hoffer2019augment} as a promising strategy for large-batch training, and has recently gained popularity when training Vision Transformers \citep{touvron2020training, touvron2021going}. The key insight of augmentation multiplicity is that we can preserve the bias in the per-example gradients introduced by data augmentation, while simultaneously suppressing the variance, by drawing multiple augmentation samples of each unique image in the batch. One can achieve this either by allowing the batch size to grow as the augmentation multiplicity increases, or by holding the batch size fixed and reducing the number of unique examples in each batch. Note that in the latter case the variance arising from data augmentation falls but the variance arising from sub-sampling the dataset increases. We study both schemes in this work. Since SGD exhibits different behaviour at different batch sizes \citep{goyal2017accurate,ma2017power,zhang2019algorithmic, smith2020generalization}, we also study augmentation multiplicity in both the small and large batch regimes. Our key findings are as follows: 
\begin{figure}[t]
	\centering
	\vskip - 3mm
    \includegraphics[width=0.9\linewidth]{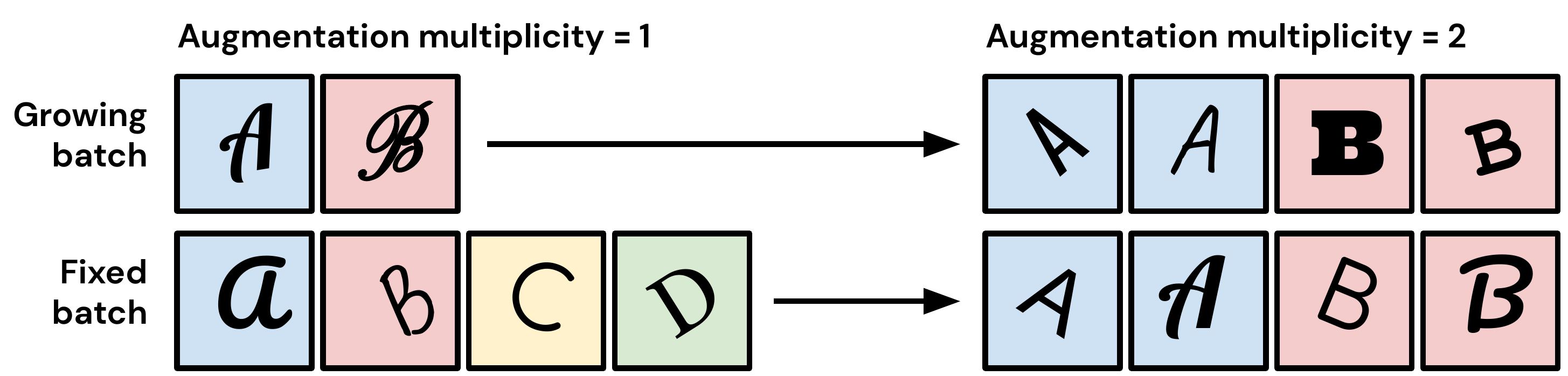}
    \vskip - 1mm
	\caption{Two forms of augmentation multiplicity. In the top row we draw 2 augmentation samples per image, while preserving the number of unique examples in the batch (batch size grows as augmentation multiplicity rises). In the bottom row we draw 2 augmentation samples per image, but hold the batch size fixed.}
	\vskip -2mm
	\label{fig:sketch}
\end{figure}
\begin{enumerate}
    \item If the number of unique images per batch is fixed (batch size grows as augmentation multiplicity increases), then test accuracy rises as augmentation multiplicity rises. This phenomenon arises even after we tune the epoch budget, implying the benefits of data augmentation arise from the bias in the gradient, not the variance. Note, by contrast, performance degrades for very large batch sizes in standard pipelines \citep{smith2020generalization}.
    
    \item Augmentation multiplicities greater than 1 also achieve higher test accuracy if we hold the batch size fixed (implying that the number of unique images per batch decreases). This benefit is particularly pronounced for large batch training, but it also arises with small batch sizes. However in this setting, despite achieving higher accuracy on the test set, large augmentation multiplicities achieve slower convergence on the training set. 

    \item To confirm our insights scale to high-performance networks, we provide an empirical evaluation of augmentation multiplicity on NFNets \citep{brock2021high}. Using an NFNet-F5 with SAM regularization and augmentation multiplicity 16, we achieve 86.8$\%$ top-1 w/o extra data on ImageNet. We achieve this result after training for 34 epochs (168900 steps), while training the same model w/o augmentation multiplicity requires significantly more epochs and reaches lower accuracy.\footnote{For clarity, we define an epoch as a single pass through the entire training set. This requires more gradient evaluations as the augmentation multiplicity increases. However we also verify that NFNets trained with large augmentation multiplicities achieve higher validation accuracy while requiring less overall compute.} 
    
    \item These observations are counter-intuitive, as when the batch size is fixed, large augmentation multiplicities increase the overall variance in the minibatch gradient estimate (since the number of unique images in each batch is reduced). However because we draw multiple samples of the augmentation procedure for each image, the variance in the per-example gradients of specific examples is reduced. We conclude that the variance in the gradient estimate arising from sub-sampling the dataset has an implicit regularization benefit enhancing generalization, but that the variance which arises from the data augmentation procedure harms test accuracy.

\end{enumerate}

As stated above, augmentation multiplicity was first proposed by \citet{hoffer2019augment} as a promising strategy for large batch training (i.e. by increasing the batch size as the augmentation multiplicity increases), while \citet{berman2019multigrain} found it can also enhance performance when the batch size is fixed. In addition, \citet{choi2019faster} explored using augmentation multiplicity to reduce communication overheads on device, and \citet{hendrycks2019augmix} showed that it can improve robustness to distributional shift. Finally, augmentation multiplicity was recently used by \citet{touvron2020training, touvron2021going} to enhance the performance of Vision Transformers \citep{dosovitskiy2020image}. However, although augmentation multiplicity has been explored in a number of works, there has been no conclusive study demonstrating that its benefits are robust after careful tuning of the compute budget and the learning rate. In addition, prior work has not proposed that the benefits of augmentation multiplicity are connected to the implicit regularization benefit of SGD \citep{keskar2016large, li2019towards, smith2021origin}.

We provide the first systematic study of how augmentation multiplicity influences deep networks trained with SGD, and we seek to explain why large augmentation multiplicities enhance generalization. We find that large augmentation multiplicities achieve higher test accuracy for both small and large batch sizes, and crucially they often achieve higher test accuracy without requiring larger compute budgets. We therefore believe large augmentation multiplicities should become the default setting in many vision applications. 

This paper is structured as follows. In Section 2, we evaluate the performance of augmentation multiplicity when the number of unique images per batch is fixed. This section demonstrates that the variance introduced by the augmentation procedure harms test accuracy. In Section 3 we evaluate the performance of augmentation multiplicity when the total batch size is fixed, in order to demonstrate that augmentation multiplicity is a practical method for enhancing model performance in standard pipelines. Next, we explore why augmentation multiplicity enhances generalization in Section 4. Finally, we apply augmentation multiplicity to the NFNet model family \citep{brock2021high} in Section 5.

\section{The benefits of data augmentation arise from bias, not variance}
\label{sec:growing_batch}
In this section, we show that the benefits of data augmentation in ResNets arise from the bias in the augmentation procedure, not the variance. We first consider a 16-4 Wide-ResNet \citep{zagoruyko2016wide}, trained on CIFAR-100 \citep{krizhevsky2009learning} with padding, left-right flips and random crops. The data augmentation process introduces a bias in the evaluated gradients. More specifically, if $\ell(x)$ denotes the loss for input $x$ and $F(x, \xi)$ denotes the augmentation function whose output is the augmented image, with $\xi$ being the collection of random variables denoting the noise in the augmentation process, then data augmentation changes the expected value of the loss since:
\begin{align}
    \mathbb{E}_\xi [\ell(F(x, \xi))] \neq \ell(x).
\end{align}
However, typically practitioners only augment each image in a minibatch once by sampling a single $\xi_1$ and computing $\ell(F(x, \xi_1))$. This introduces variance into the training process. To disentangle the effects of bias and variance on final model performance, we train with a range of augmentation multiplicities $n$, meaning that for each unique input $x$, we produce $n$ augmented inputs by drawing $n$ random samples of the noise $\xi$, and average the loss over these $n$ samples:
\begin{align}
    \frac{1}{n} \sum_{i=1}^n \ell(F(x, \xi_i)).
\end{align}
Large augmentation multiplicities reduce the gradient covariance, helping to isolate the effect of bias. In the CIFAR-100 experiments in this section, we draw 64 unique inputs in each minibatch, such that the total batch size $B=64n$ grows with augmentation multiplicity $n$. Note that we keep the number of unique examples in the minibatch fixed in this section, allowing the batch size to grow as the augmentation multiplicity rises, in order to isolate the role of the variance which arises from the augmentation procedure, without increasing the variance from sub-sampling the dataset.
% In this section, we demonstrate that the primary benefits of data augmentation in ResNets arise from the bias in the augmentation procedure and not the variance. To establish this, we consider a 16-4 Wide-ResNet \citep{zagoruyko2016wide}, trained on CIFAR-100 with padding, left-right flips and random crops. In all the experiments shown in this section, we draw 64 unique inputs in each minibatch. However, we train with a range of augmentation multiplicities $n$, meaning that we draw $n$ random samples from the data augmentation procedure for each unique input, such that the total batch size $B=64n$ grows as the augmentation multiplicity increases. Consequently the variance introduced by the data augmentation operation is suppressed as the augmentation multiplicity rises.
\begin{figure}[t]
\centering
  \vskip -3mm
\subfigure[]{\includegraphics[height=3.0cm]{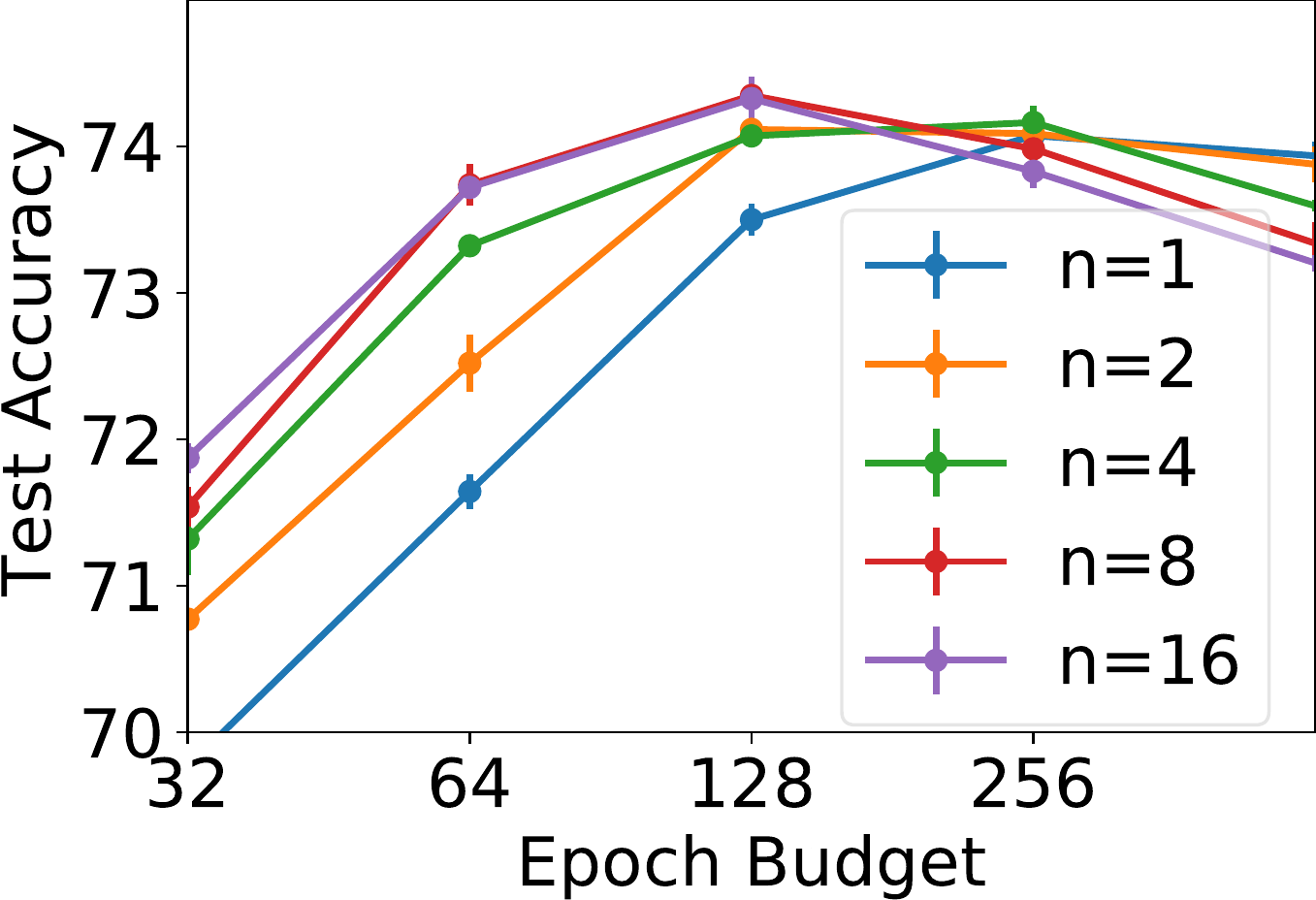}\label{fig:growing_a}}
\subfigure[]{\includegraphics[height=3.0cm]{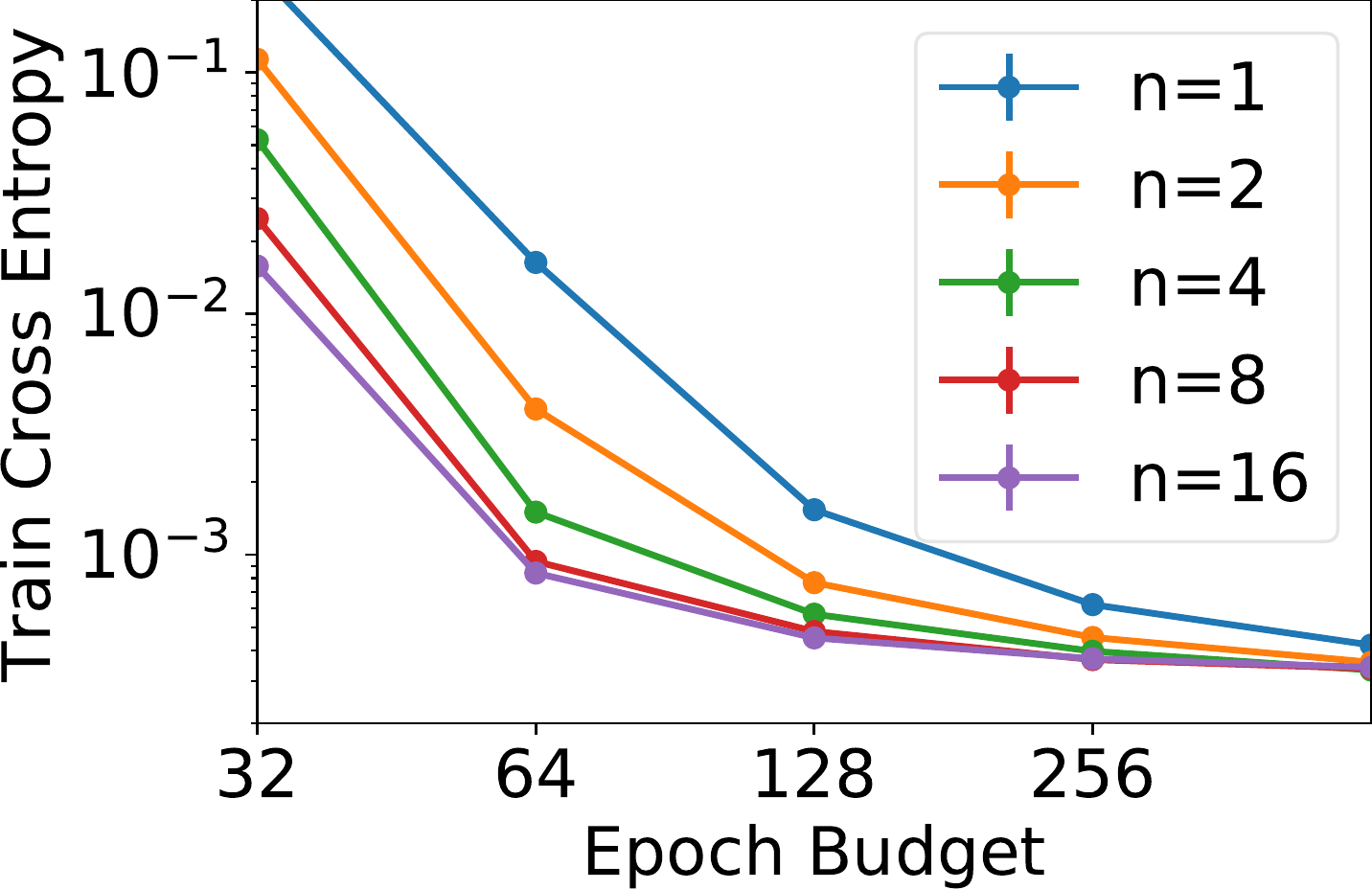}\label{fig:growing_b}}
\subfigure[]{\includegraphics[height=3.0cm]{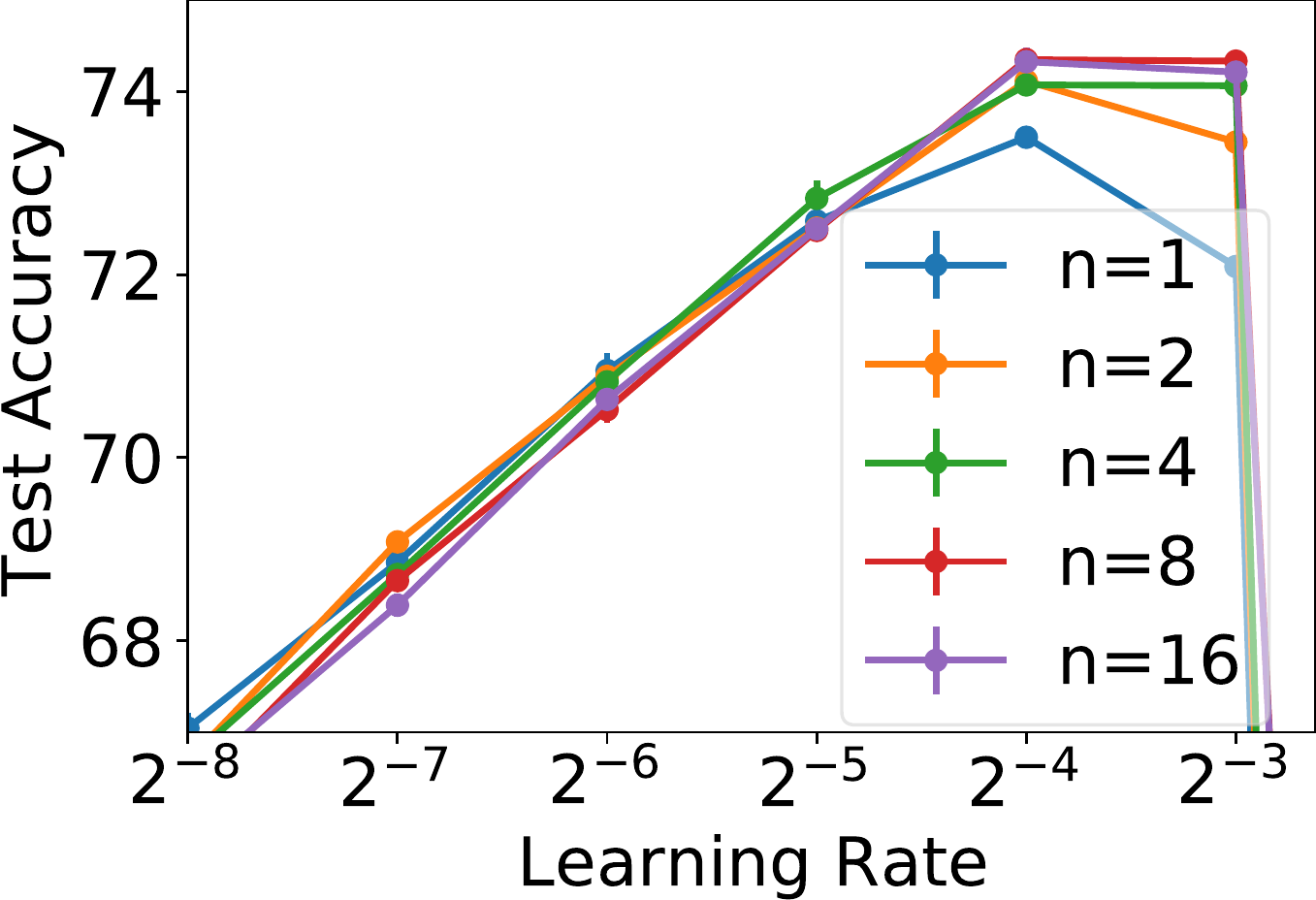}\label{fig:growing_c}}
  \vskip -2.5mm
\caption{Results with a 16-4 Wide-ResNet on CIFAR-100. The number of unique images per batch is fixed, such that the batch size grows as the augmentation multiplicity $n$ increases. (a) Test accuracies at optimal learning rates for a range of epoch budgets. Large augmentation multiplicities achieve higher test accuracy and require fewer training epochs. (b) The training cross entropy for a range of epoch budgets.
% plotted at the learning rate in our grid which minimizes the cross entropy. The cross entropy is evaluated on the full training dataset using the raw images (without data augmentation). 
In this growing batch setting, large augmentation multiplicities achieve faster convergence. (c) The test accuracy at a range of learning rates for a compute budget of 128 epochs. Different augmentation multiplicities achieve similar test accuracy for small learning rates, but large augmentation multiplicities achieve higher test accuracy with large learning rates.
%\razp{Is the training loss computed on the same data ?! for fairness .. and is this with augmentation or without? or is this just the training error being minimized -> answered in the text }
}
  \vskip -2.5mm
\label{fig:growing_cifar}
\end{figure}

The performance of batch-normalized networks depends strongly on the examples used to evaluate the batch statistics. Therefore to simplify our analysis, we train our Wide-ResNets without normalization using the SkipInit initialization scheme \citep{de2020batch}. We use SGD with a momentum coefficient of $0.9$ and weight decay with a coefficient of $5\times 10^{-4}$. When training for $m$ epochs, the learning rate is constant for the first $m/2$ epochs, and then decays by a factor of 2 every remaining $\frac{m}{20}$ epochs. We provide the mean accuracy of the best 5 out of 7 training runs. %and we provide additional experiments with batch normalization in appendix \ref{app:BN}, for which we observe similar behaviour.

In Figure \ref{fig:growing_a}, we plot the test accuracy for a range of augmentation multiplicities, across a range of epoch budgets. We independently tune the learning rate on a logarithmic grid spaced by factors of 2 for each combination of epoch budget and augmentation multiplicity. After tuning the epoch budget, we find that large augmentation multiplicities achieve slightly higher test accuracy. For instance, augmentation multiplicity 16 achieves a peak test accuracy of $74.3 \pm 0.1\%$ after 128 epochs, while augmentation multiplicity 1 (normal training) achieves a peak of $74.1 \pm 0.01\%$ after 256 epochs. Since higher augmentation multiplicities reduce the variance from data augmentation without changing the bias, this confirms that the variance arising from data augmentation reduces the test accuracy. Large augmentation multiplicities also require fewer training epochs/parameter updates to reach optimal performance, indicating that the variance from data augmentation slows down training. %\footnote{Intriguingly, large augmentation multiplicities do achieve lower test accuracy for very large epoch budgets (larger than optimal), suggesting that the variance from the augmentation procedure can reduce over-fitting when models are over-trained.} 
For completeness, we note that training without data augmentation at batch size 64 achieves a substantially lower test accuracy of 61.4$\%$ after tuning the learning rate and compute budget, verifying that, as expected, the bias introduced by data augmentation significantly enhances generalization. 
Note that a similar experiment appeared in \citet{hoffer2019augment}, however our study is the first to retune the learning rate and epoch budget for each augmentation multiplicity. This is essential to establish whether the variance from data augmentation is detrimental by eliminating the possibility that the original hyper-parameters were sub-optimal.

In Figure \ref{fig:growing_b}, we demonstrate that large augmentation multiplicities also achieve faster convergence on the training set (per epoch/per parameter update) in this ``growing batch'' setting. This is intuitive, since large multiplicities perform more gradient evaluations per minibatch, reducing the variance of the gradient estimate. For clarity, we plot the training loss at the learning rate in our grid which minimizes training loss. We evaluate the training loss on the full training set using the raw images (without data augmentation). 

Finally in Figure \ref{fig:growing_c}, we plot how the test accuracy for a budget of 128 epochs depends on the learning rate. Note that 128 epochs is the optimal epoch budget for large augmentation multiplicities. %Note that in this setting, large augmentation multiplicities perform more gradient evaluations but the same number of parameter updates. 
We find that different multiplicities achieve similar test accuracy when the learning rate is small, but large augmentation multiplicities achieve higher test accuracy when the learning rate is large, and this enables large augmentation multiplicities to achieve higher overall test accuracy after tuning. These results suggest that the variance in the data augmentation operation reduces the stability of training, which impairs training with large learning rates and consequently reduces test performance. We discuss how large learning rates benefit generalization in more detail in Section \ref{sec:analysis} \citep{li2019towards, smith2021origin}.
\begin{figure}[t]
\centering
  \vskip -3mm
\subfigure[]{\includegraphics[height=3.0cm]{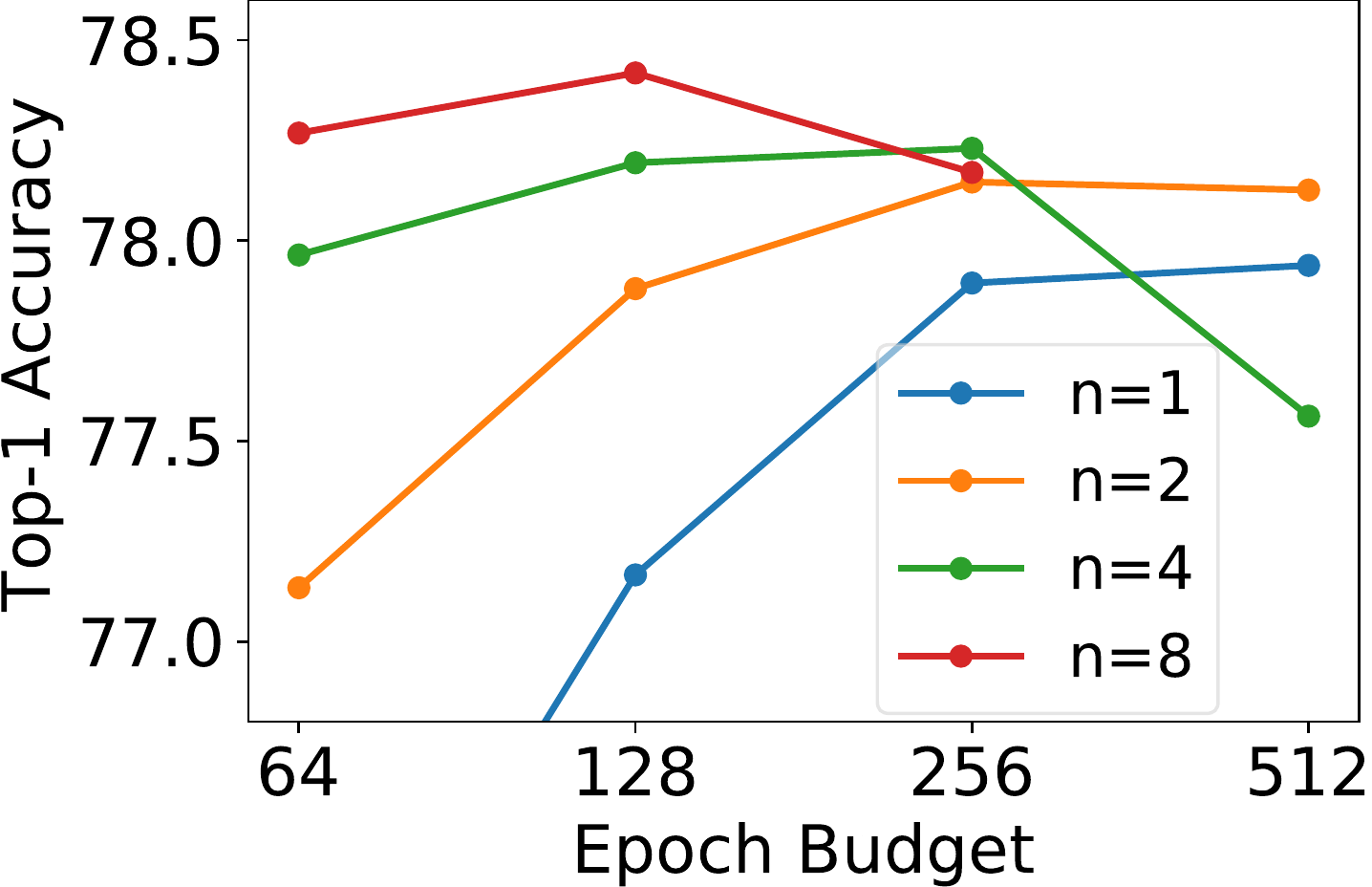}\label{fig:growing_rn50_a}}
\subfigure[]{\includegraphics[height=3.0cm]{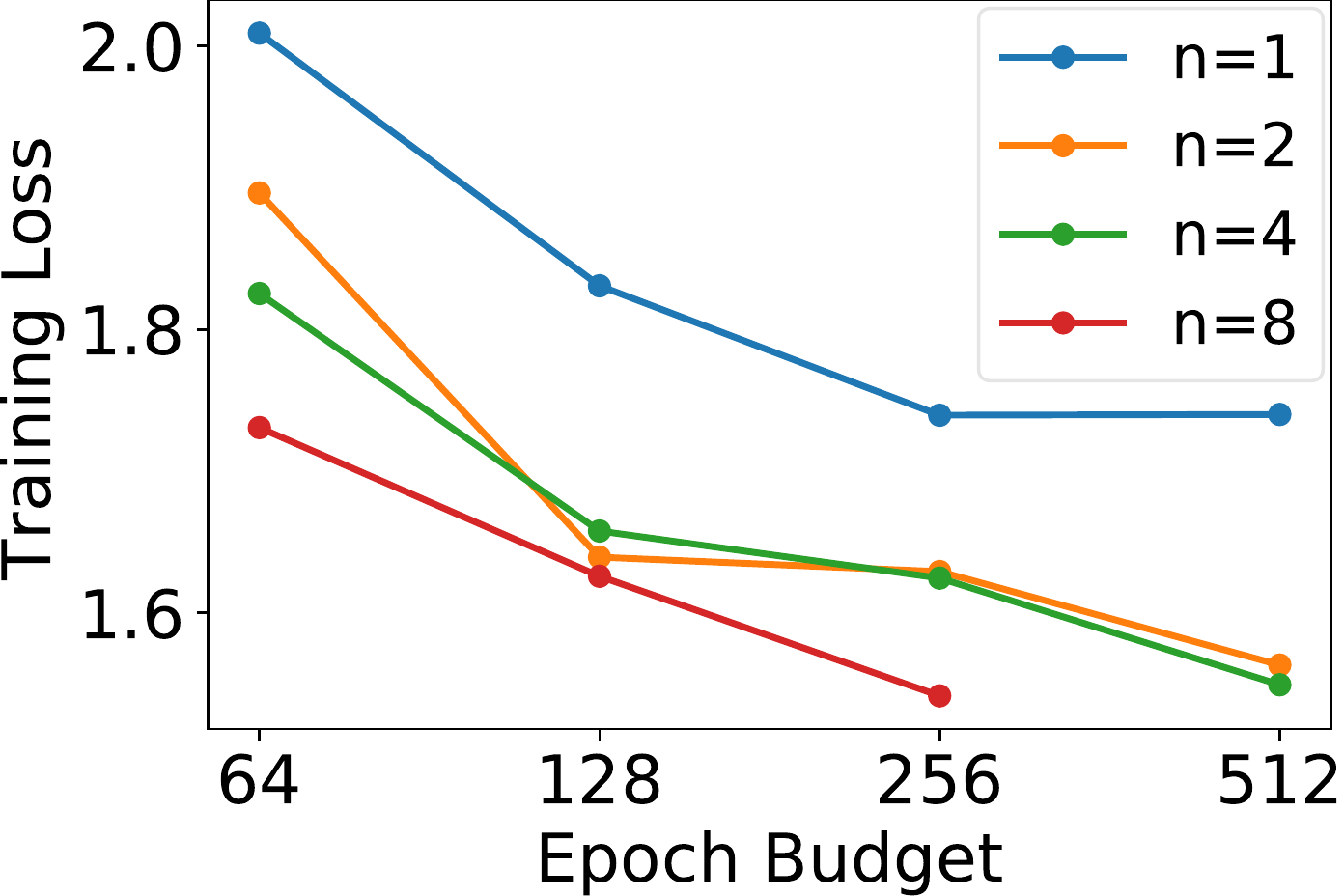}\label{fig:growing_rn50_b}}
\subfigure[]{\includegraphics[height=3.0cm]{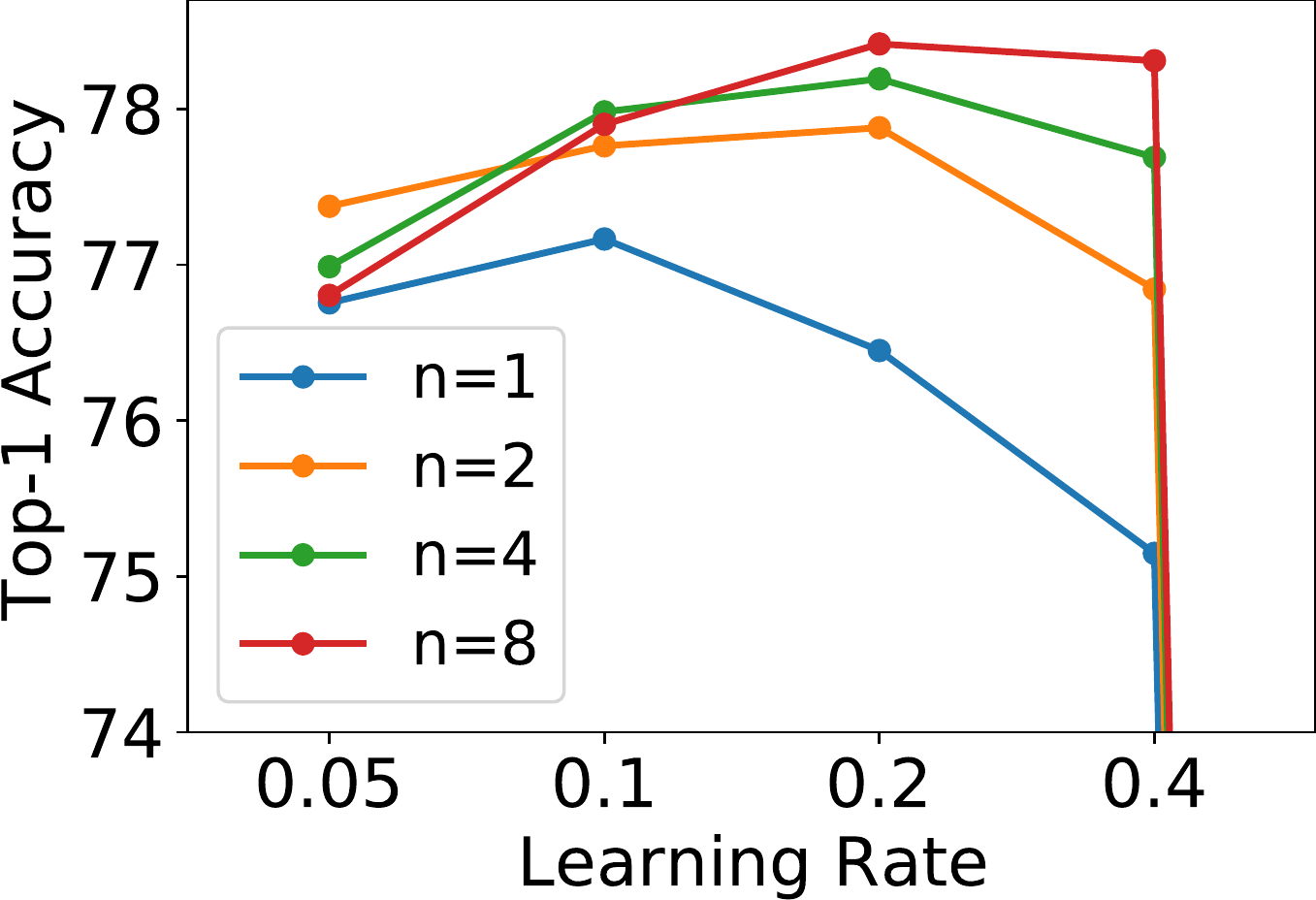}\label{fig:growing_rn50_c}}
  \vskip -2.5mm
\caption{Results with an NF-ResNet-50 on ImageNet. The number of unique images per batch is fixed, such that the batch size grows as the augmentation multiplicity $n$ increases. (a) Top-1 accuracy at optimal learning rates for a range of epoch budgets. Larger augmentation multiplicities achieve higher test accuracies and also require fewer training epochs.
% Large augmentation multiplicities achieve higher overall test accuracies, and also require fewer training epochs. 
(b) The training cross entropy for a range of epoch budgets. Note that we evaluate the cross entropy on a single minibatch, not the full dataset.
% In this growing batch setting, large augmentation multiplicities achieve higher training accuracy. 
% \vskip -1mm
(c) The test accuracy at a range of learning rates for a compute budget of 128 epochs.
% Different multiplicities achieve similar test accuracy when the learning rate is small, but large multiplicities achieve higher test accuracy with large learning rates.
%Similar to Figure \ref{fig:growing_cifar}, larger augmentation multiplicities achieve higher test accuracies.
%\razp{Is the training loss computed on the same data ?! for fairness .. and is this with augmentation or without? or is this just the training error being minimized -> answered in the text }
}
\label{fig:growing:imagenet}
  \vskip -2.5mm
\end{figure}

We provide additional results in Figure \ref{fig:growing:imagenet}, for a ResNet-50 \citep{he2016identity} trained on ImageNet \citep{ILSVRC15} with 256 unique images per batch using the Normalizer-Free (NF-) strategy of \citet{brock2021characterizing}. We use SGD with Momentum coefficient 0.9,  cosine learning rate decay \citep{loshchilov2016sgdr} and baseline pre-processing including left-right flips and random crops \citep{szegedy2017inception}. 
% We also apply stochastic depth with a drop rate of 0.1 \citep{huang2016deep} and dropout with a drop rate of 0.25 \citep{srivastava2014dropout}. 
Following \citet{brock2021characterizing}, we use weight decay with a coefficient of $5\times10^{-5}$, label smoothing with a coefficient of 0.1 \citep{szegedy2016rethinking}, stochastic depth with a drop rate of 0.1 \citep{huang2016deep} and dropout before the final linear layer with a drop probability of 0.25 \citep{srivastava2014dropout}. 
We tune the learning rate on a logarithmic grid spaced by factors of 2, performing a single training run at each learning rate. Once again, we observe that large augmentation multiplicities achieve higher top-1 accuracies while also requiring fewer training epochs. For instance, augmentation multiplicity 1 (normal training) achieves a peak top-1 accuracy of 77.9$\%$ after 512 epochs, while augmentation multiplicity 8 achieves a peak top-1 accuracy of 78.4$\%$ after 128 epochs.

\section{An empirical evaluation of augmentation multiplicity for fixed batch sizes}

\label{sec:fixed}

In Section \ref{sec:growing_batch}, we established that the generalization benefit of data augmentation arises from the bias it introduces into the gradient estimate, while the variance from data augmentation both slows down optimization and impedes generalization. We also showed that large augmentation multiplicities can achieve higher test accuracy even after tuning the compute budget. However we allowed the minibatch size to grow as the augmentation multiplicity rose. This is impractical, since the batch size is usually determined by the hardware available for training. In this section, we confirm that large augmentation multiplicities continue to achieve superior test accuracy if we maintain a constant batch size $B$, such that the number of unique training examples per minibatch declines as the augmentation multiplicity increases. The behaviour of SGD differs substantially in the small batch, `noise dominated' regime and the large batch, `curvature dominated' regime \citep{ma2017power,zhang2019algorithmic, smith2020generalization}. We consider both regimes here.

%We consider two models; A 16-4 Wide-ResNet trained on CIFAR-100 at batch sizes 64 (small) and 1024 (large), and an NF-ResNet-50 \citep{brock2021characterizing} trained on ImageNet at batch sizes 256 (small) and 4096 (large). We described the Wide-ResNet model and CIFAR-100 training pipeline in section \ref{sec:growing_batch}. We train the NF-ResNet50 using SGD with Momentum coefficient 0.9. We use cosine learning rate decay \citep{loshchilov2016sgdr} and baseline pre-processing including flips and random crops \citep{szegedy2017inception}. We also apply stochastic depth with a drop rate of 0.1 \citep{huang2016deep} and dropout with a drop rate of 0.25 \citep{srivastava2014dropout}. We tune the learning rate on a logarithmic grid spaced by factors of 2, performing a single training run at each learning rate.

\begin{figure}[t]
\centering
 \vskip -1mm
\subfigure[CIFAR-100]{\includegraphics[width=3.43cm]{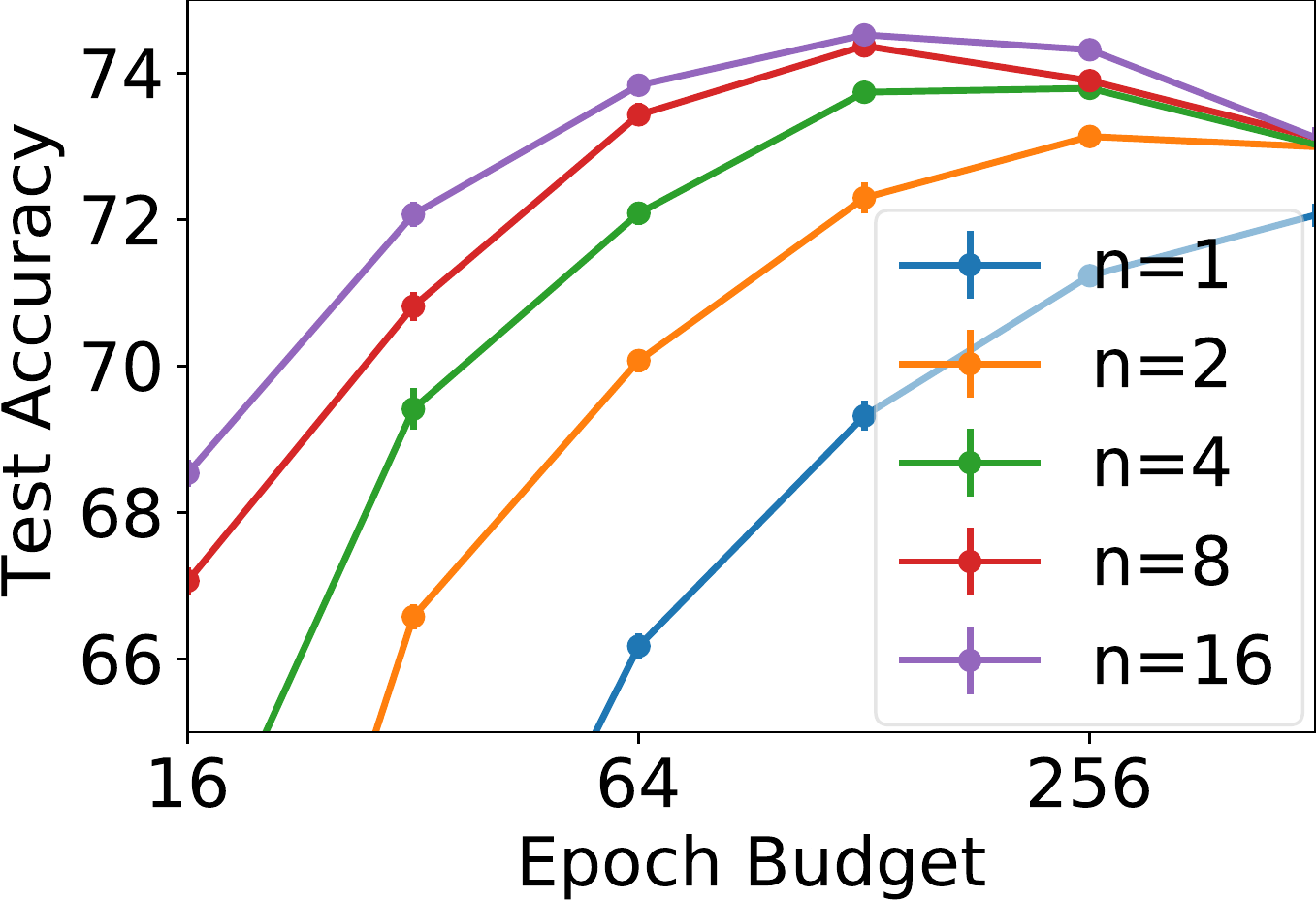}\label{fig:large_a}}
\subfigure[CIFAR-100]{\includegraphics[width=3.43cm]{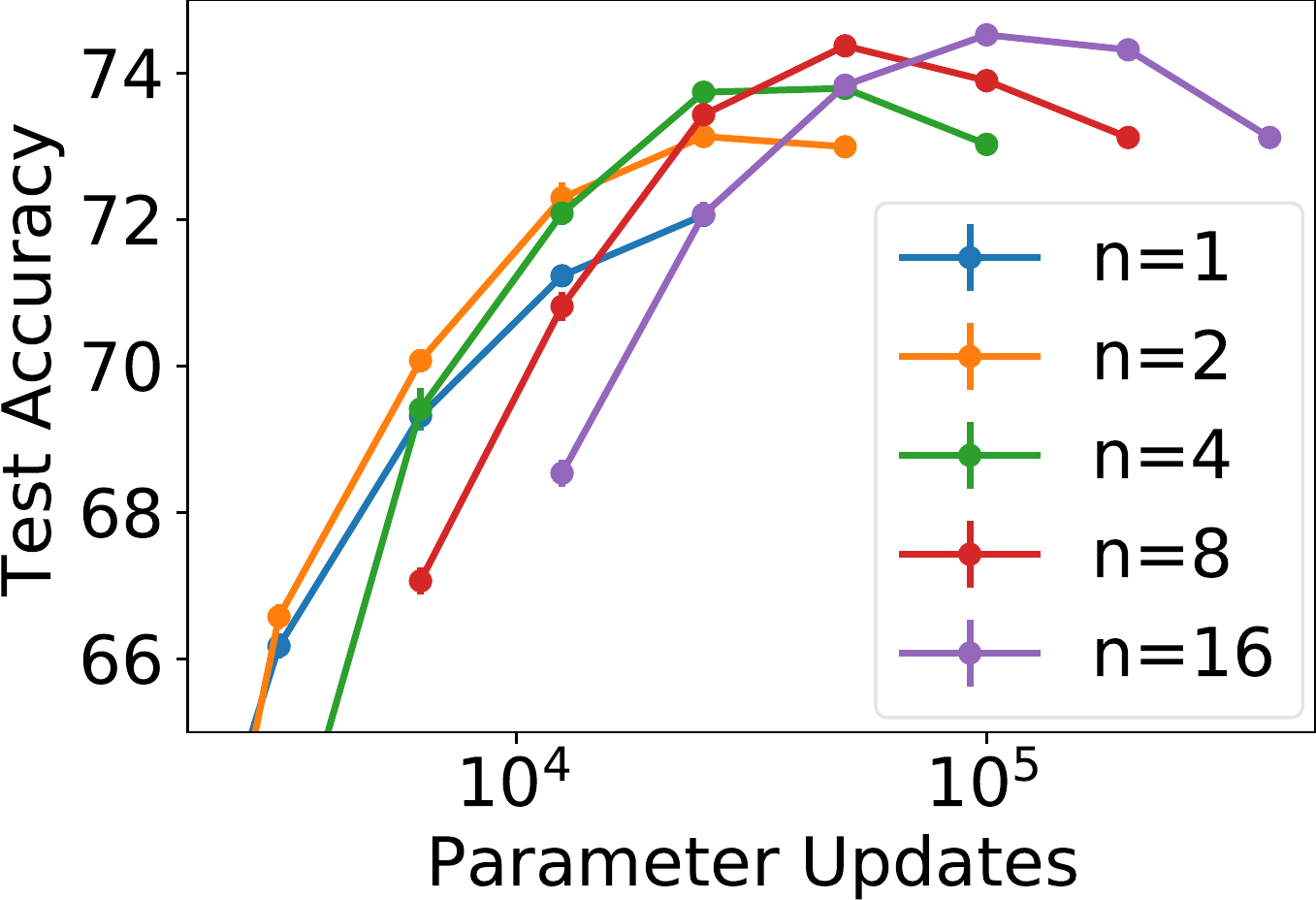}\label{fig:large_b}}
\subfigure[ImageNet]{\includegraphics[width=3.43cm]{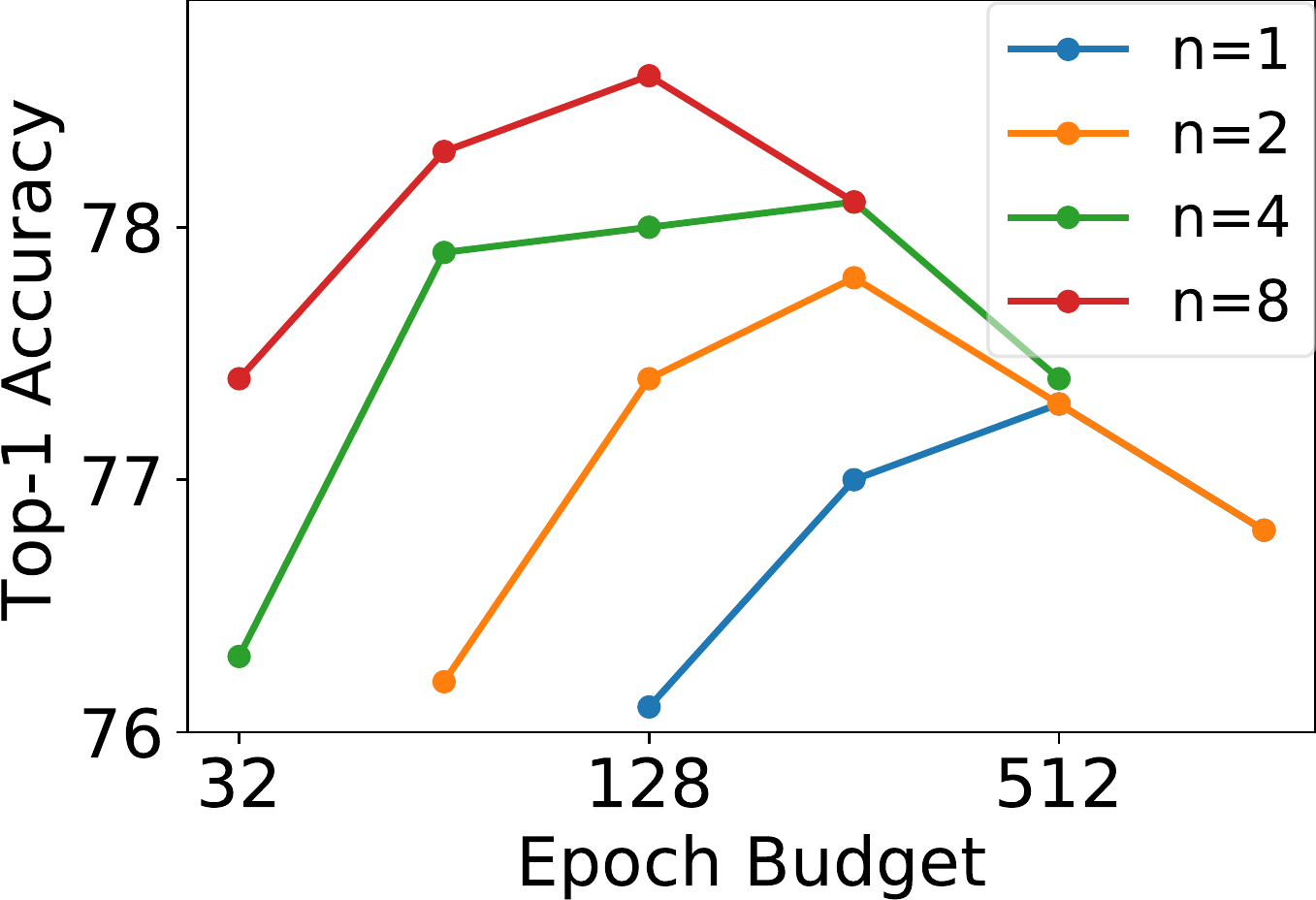}\label{fig:large_c}}
\subfigure[ImageNet]{\includegraphics[width=3.43cm]{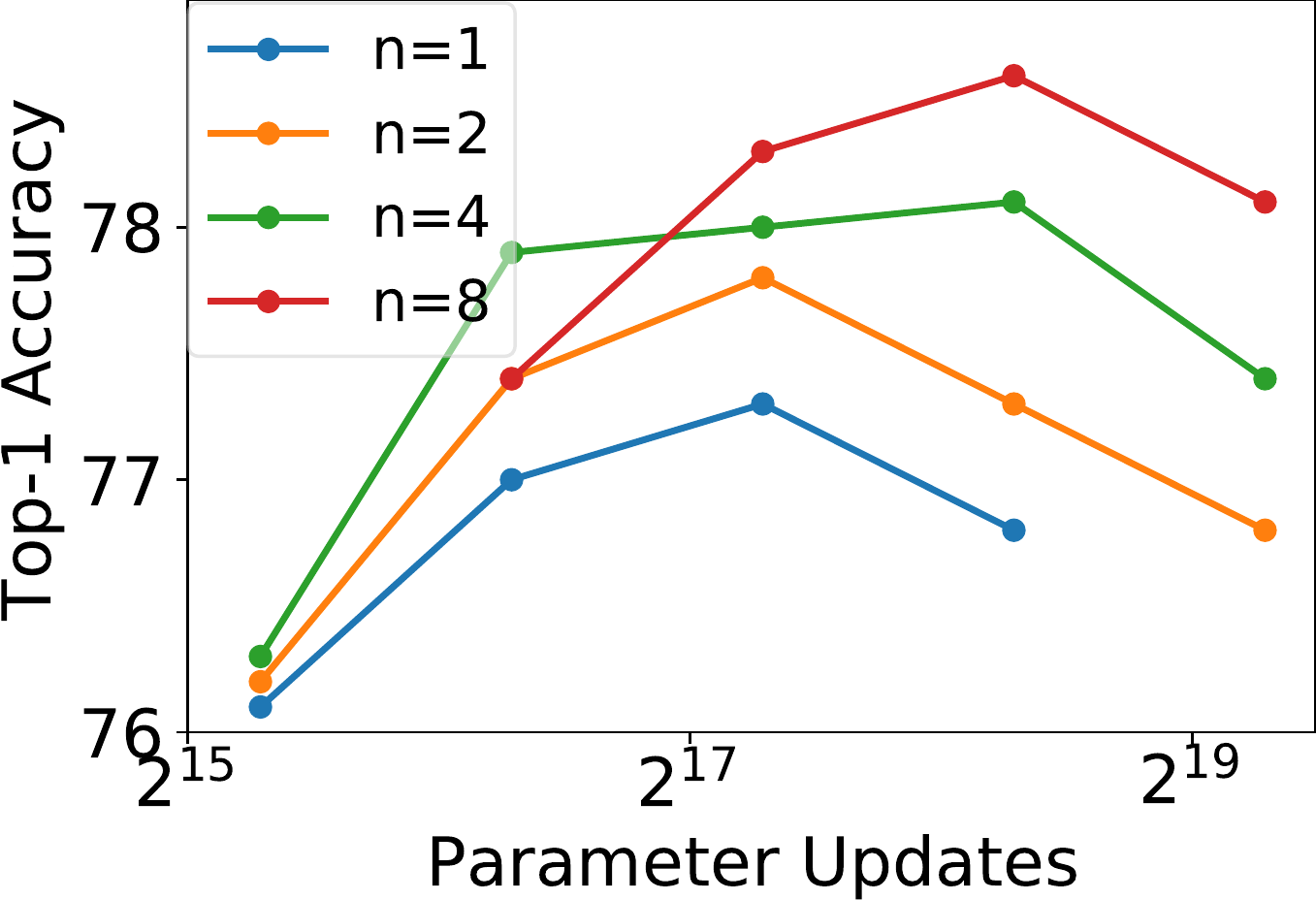}\label{fig:large_d}}%
\vskip -2mm
\caption{Large augmentation multiplicities achieve significantly higher test accuracy after large batch training. Here the batch size is independent of the augmentation multiplicity n, such that the number of unique images per batch declines as augmentation multiplicity increases. We emphasize that large multiplicities perform more parameter updates per epoch here. We show the test accuracy achieved for a given epoch budget (a/c) and for a given number of parameter updates (b/d). Figures a and b consider a 16-4 Wide-ResNet trained on CIFAR-100 at batch size 1024, while figures c and d consider an NF-ResNet50 trained on ImageNet at batch size 4096. In both cases, large augmentation multiplicities achieve substantially higher test accuracy, while also requiring fewer training epochs and without requiring more parameter updates.
}
\label{fig:large}
\vskip -2mm
\end{figure}

As before, we consider two models; A 16-4 Wide-ResNet trained on CIFAR-100 at batch sizes 64 (small) and 1024 (large), and an NF-ResNet-50 \citep{brock2021characterizing} trained on ImageNet at batch sizes 256 (small) and 4096 (large). We follow the same training pipelines as described in Section \ref{sec:growing_batch}.
In Figure \ref{fig:large}, we plot the performance of both models in the large batch limit for a range of compute budgets. Note that since the number of unique images per batch now decreases as the augmentation multiplicity increases, larger multiplicities perform more parameter updates per epoch of training. We therefore provide the test accuracy achieved for a given epoch budget in figures a and c, as well as the test accuracy achieved within a given number of parameter updates in figures b and d. For clarity, since the batch size is fixed in this section, the computational cost of training is proportional to the number of parameter updates. Considering first figures a and c, for both models large augmentation multiplicities achieve significantly higher test accuracy while requiring a smaller epoch budget. For instance on the ResNet-50, augmentation multiplicity 8 achieves $78.6\%$ top-1 accuracy after 128 epochs, while augmentation multiplicity 1 achieves a peak accuracy of $77.3\%$ after 512 epochs. 

\begin{figure}[t]
\centering
  \vskip -1mm
\subfigure[CIFAR-100]{\includegraphics[width=3.43cm]{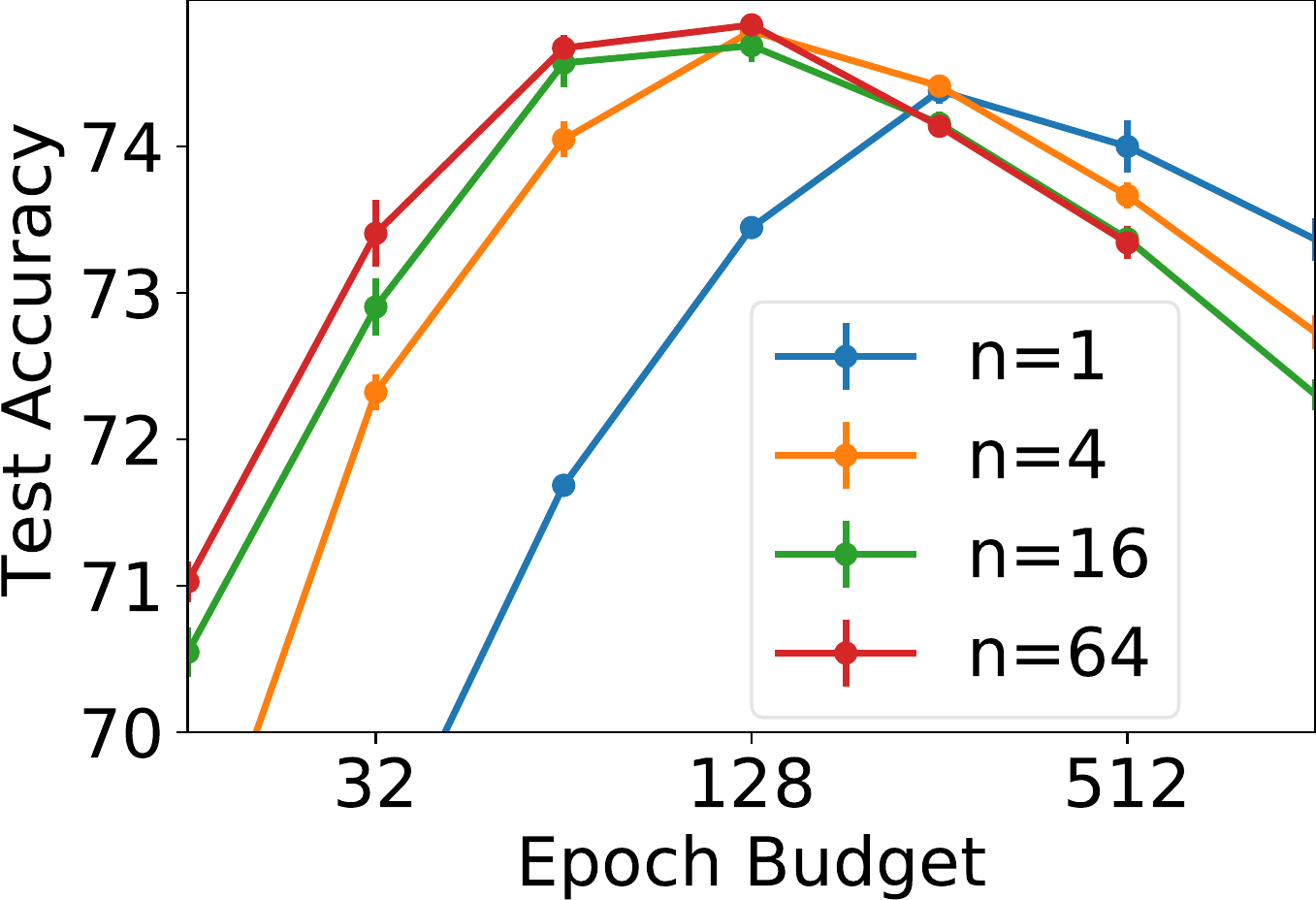}\label{fig:small_a}}
\subfigure[CIFAR-100]{\includegraphics[width=3.43cm]{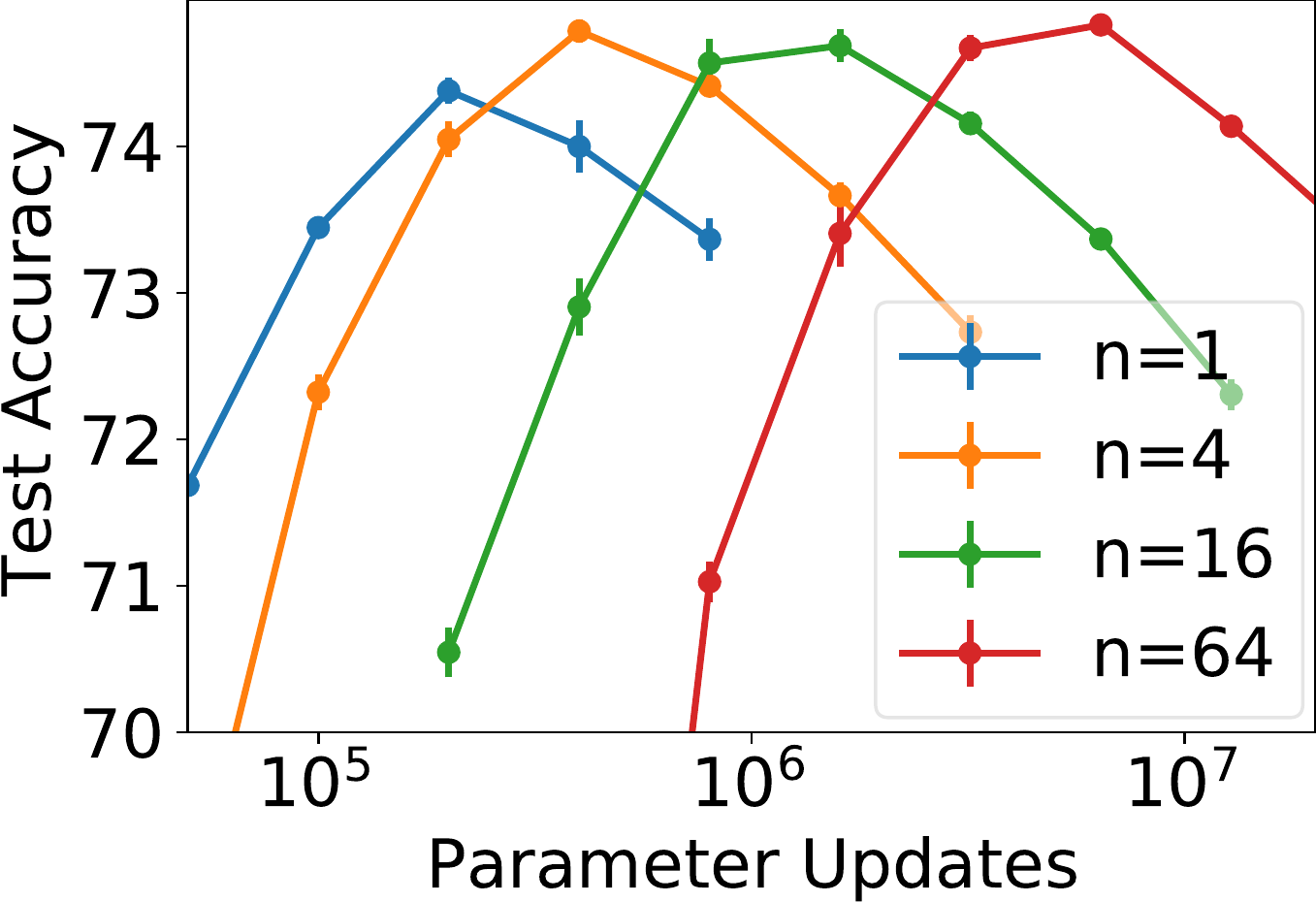}\label{fig:small_b}}
\subfigure[ImageNet]{\includegraphics[width=3.43cm]{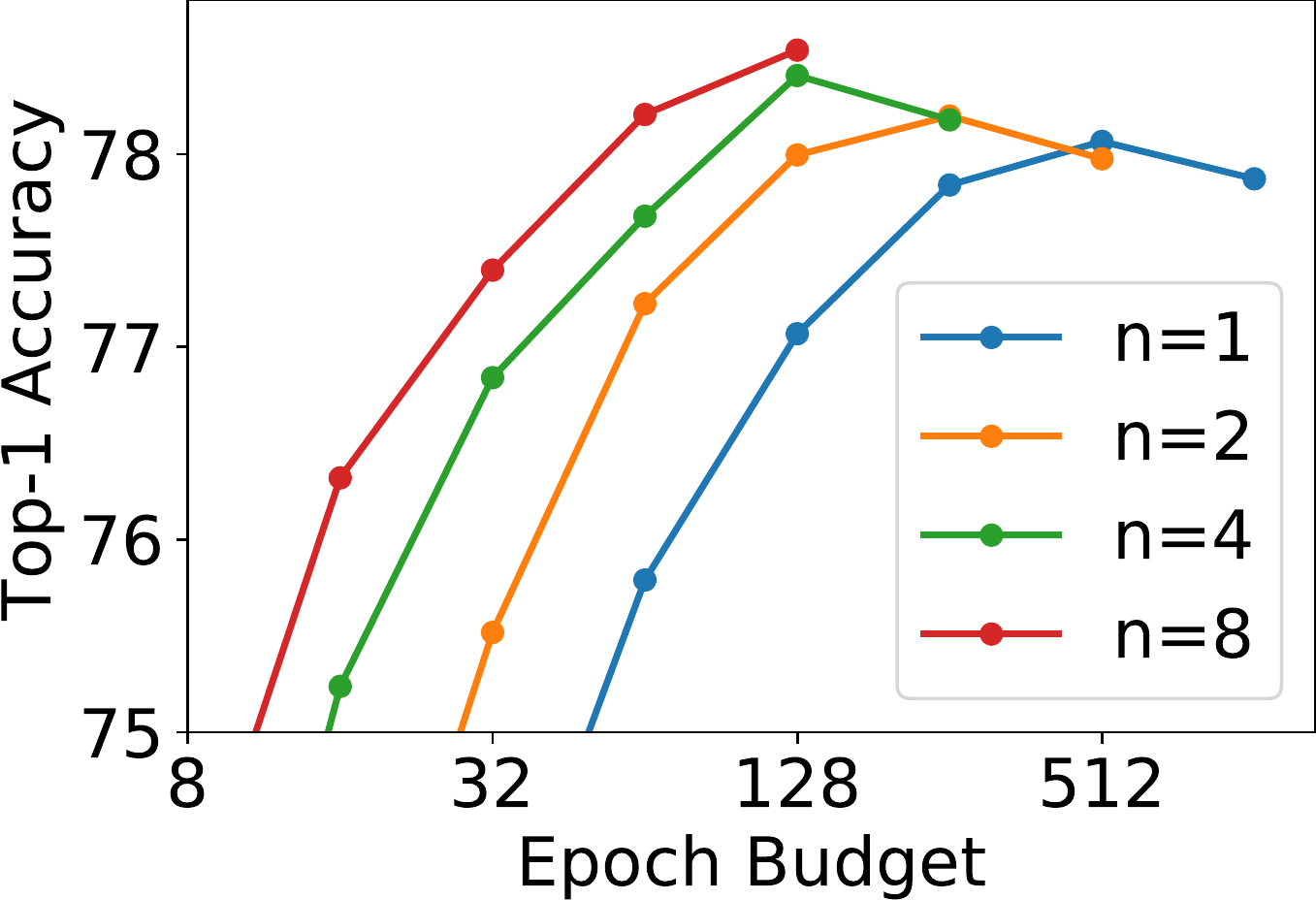}\label{fig:small_c}}
\subfigure[ImageNet]{\includegraphics[width=3.43cm]{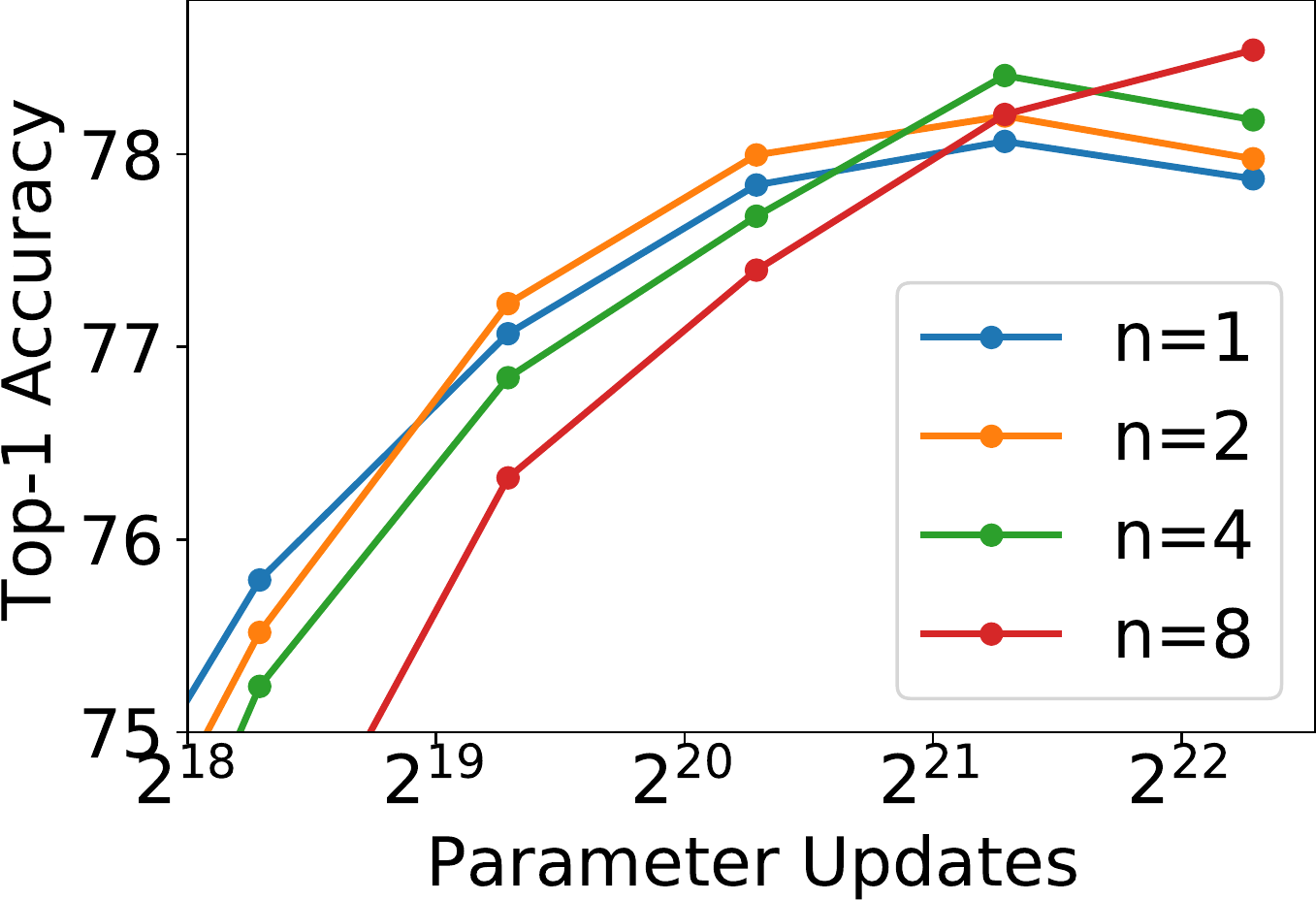}\label{fig:small_d}}%
\vskip -2mm
\caption{Large augmentation multiplicities achieve higher test accuracy for small batch training. We show the test accuracy achieved for a given epoch budget (a/c) and a given number of parameter updates (b/d). Figures a and b consider a 16-4 Wide-ResNet trained on CIFAR-100 with a batch size of 64, while figures  c and d consider an NF-ResNet50 trained on ImageNet at batch size 256. In both cases augmentation multiplicities greater than 1 achieve higher test accuracy in fewer training epochs while requiring similar or slightly larger numbers of parameter updates.
}
\label{fig:small}
\vskip -2mm
\end{figure}

In figures b and d, we observe that moderately large augmentation multiplicities also achieve higher test accuracy than normal training (n=1) without requiring more parameter updates (and therefore without requiring more compute) \citep{berman2019multigrain}. For instance on the Wide-ResNet, augmentation multiplicity 1 achieves $72.1 \pm 0.1\%$ after $4\times 10^5$ updates, while multiplicity 4 achieves $73.7 \pm 0.04\%$.

In Figure \ref{fig:small}, we provide a similar panel of figures when training the same two models in the small batch limit. Considering first figures \ref{fig:small_a} and \ref{fig:small_c}, on both CIFAR-100 and ImageNet, augmentation multiplicities larger than 1 continue to achieve higher test accuracy while also requiring fewer training epochs. On CIFAR-100, the benefits of augmentation multiplicity saturate for multiplicities $n \gtrsim 4$, while on ImageNet the peak test accuracy continues to rise for all multiplicities considered. Note however that in the small batch regime large augmentation multiplicities typically require more parameter updates (figures \ref{fig:small_b} and \ref{fig:small_d}). 

Based on the experiments presented in sections \ref{sec:growing_batch} and \ref{sec:fixed}, we conclude that augmentation multiplicity is not only beneficial during large batch training \citep{hoffer2019augment}, but that it also enhances generalization when the batch size is fixed. This phenomenon was first observed by \citet{berman2019multigrain}, however our study is the first to consider multiple batch sizes and to verify that the phenomenon is robust to retuning both the compute budget and the learning rate.

\section{Why do large augmentation multiplicities enhance generalization?}
\label{sec:analysis}

\begin{figure}[t]
\centering
 \vskip -1mm
\subfigure[Large batch]{\includegraphics[height=2.28cm]{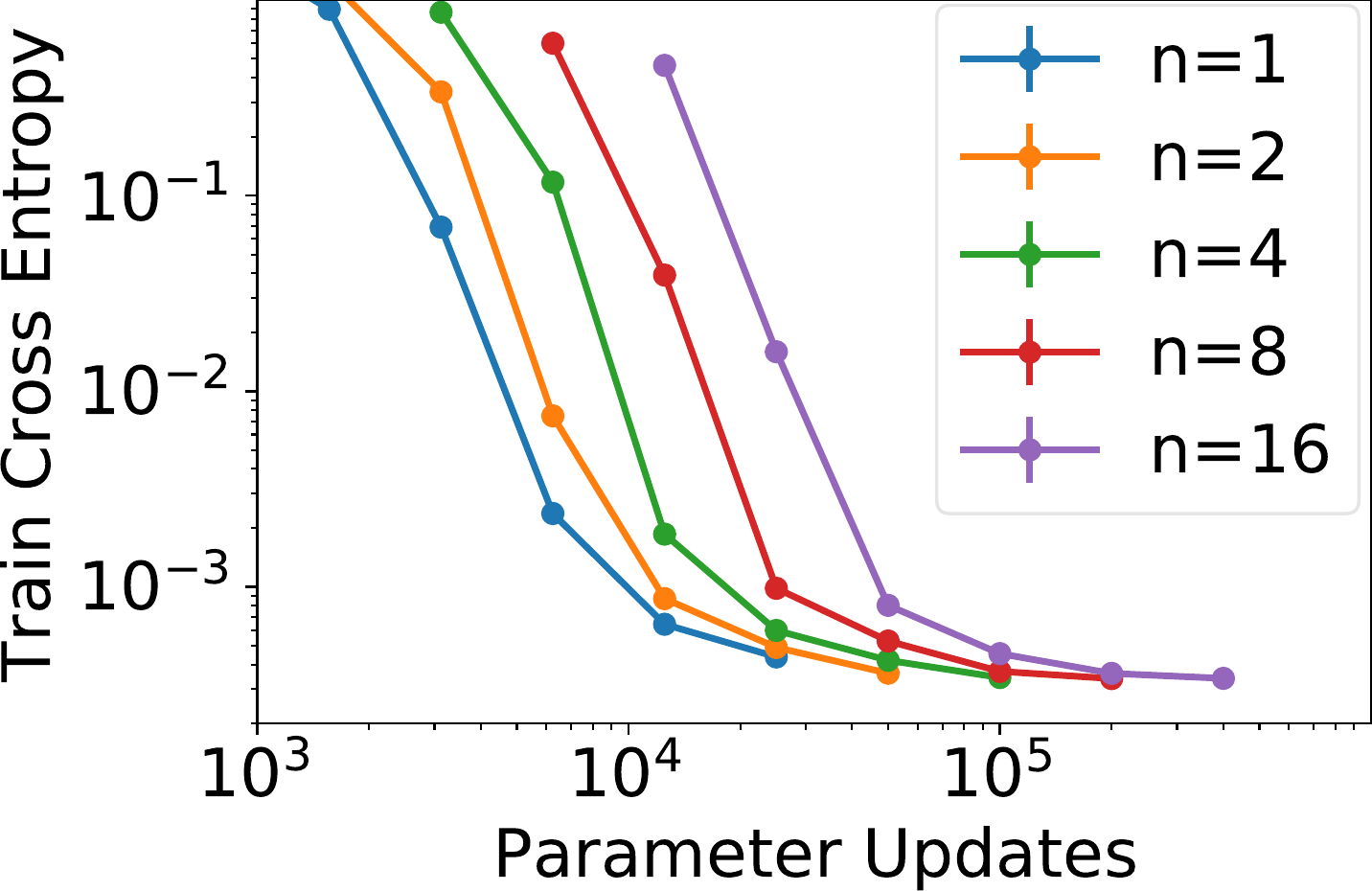}\label{fig:why:a}}
\subfigure[Small batch]{\includegraphics[height=2.28cm]{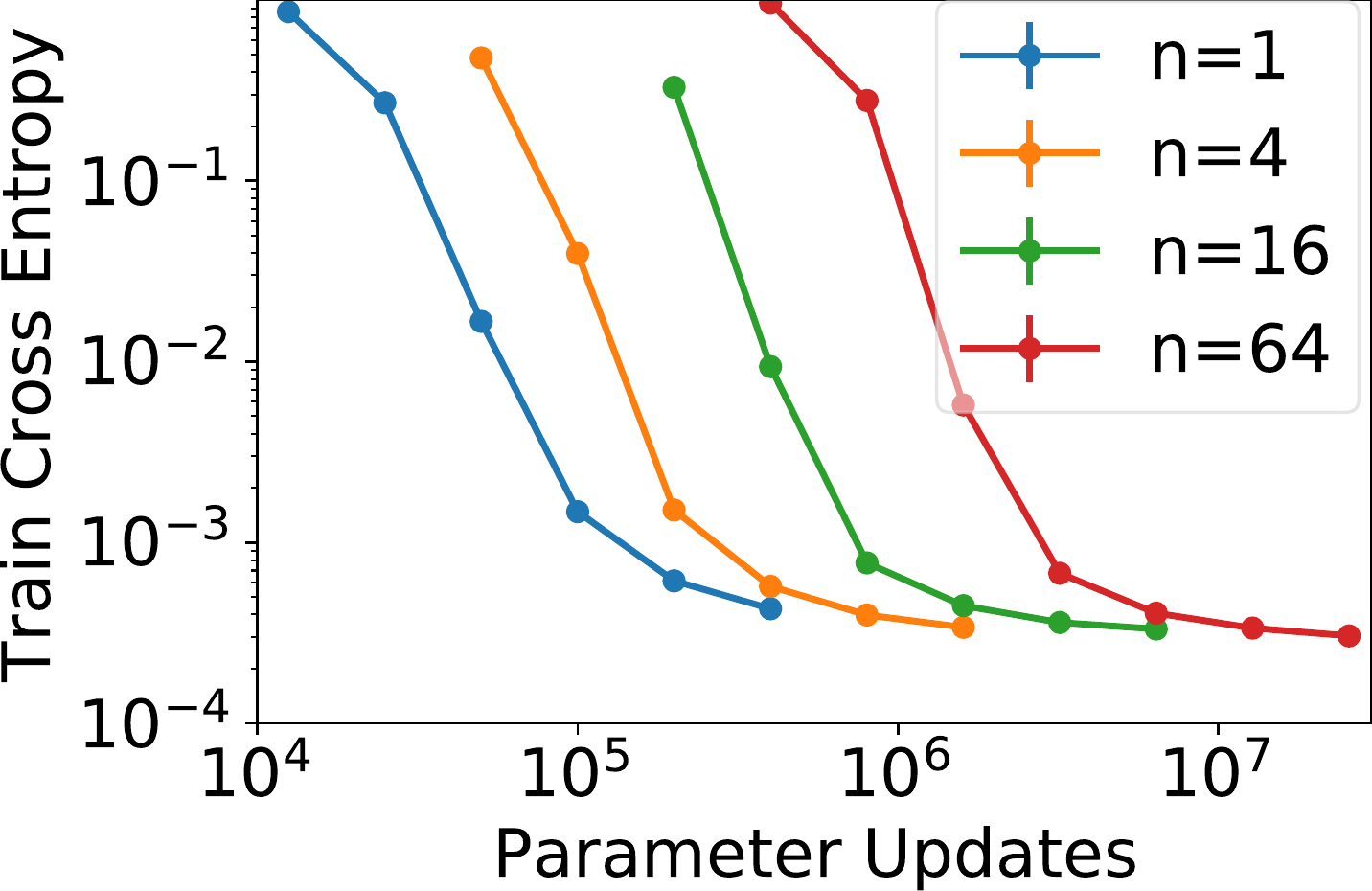}\label{fig:why:b}}
\subfigure[Large batch]{\includegraphics[height=2.28cm]{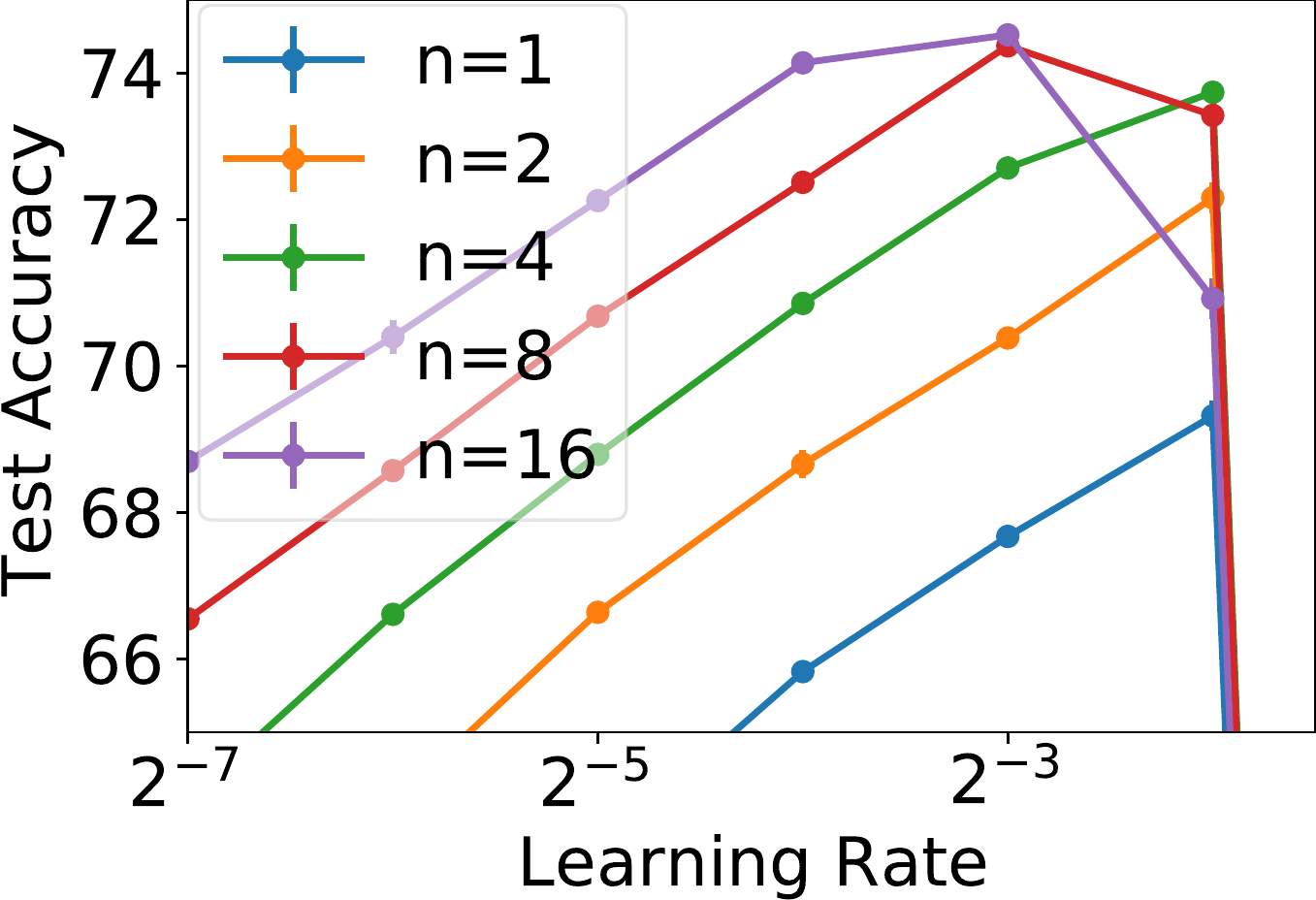}\label{fig:why:c}}
\subfigure[Small batch]{\includegraphics[height=2.28cm]{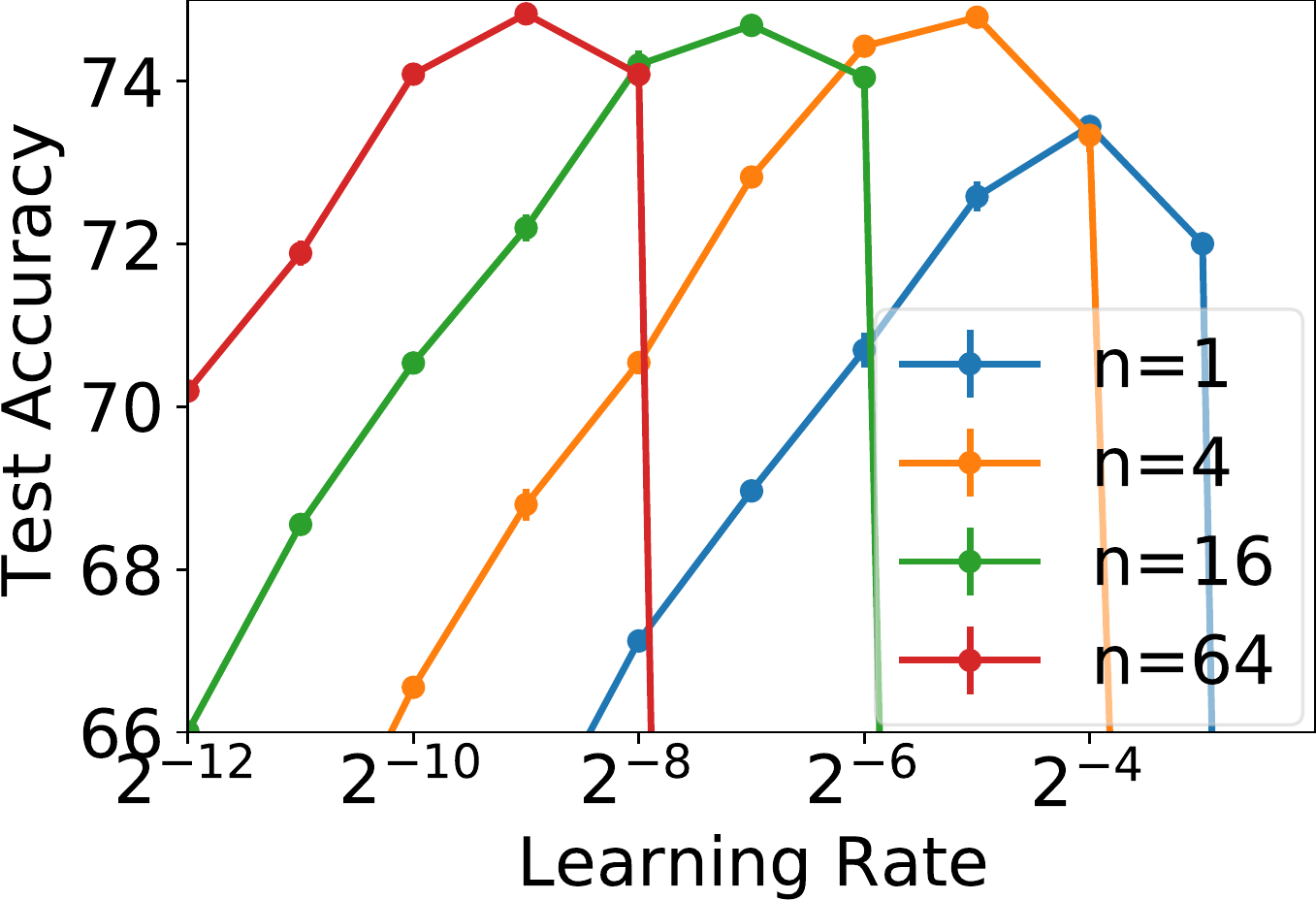}\label{fig:why:d}}
 \vskip -2mm
\caption{A 16-4 Wide-ResNet trained on CIFAR-100 at batch size 1024 and 64. Small augmentation multiplicities achieve lower training loss after a given number of parameter updates (a/b). For large batch sizes, large augmentation multiplicities achieve higher test accuracy without changing the optimal learning rate (c). For small batch sizes, large augmentation multiplicities  achieve higher test accuracy but must be trained with smaller learning rates (d). 
}
 \vskip -2mm
\end{figure}

In Section \ref{sec:fixed}, we demonstrated that augmentation multiplicities larger than 1 achieve higher test accuracy for both small and large batch training, despite containing fewer unique training examples in each minibatch. This phenomenon is highly surprising since, as observed by \citet{hoffer2019augment}, gradients evaluated on different augmentations of the same image are correlated, while gradients evaluated on independent training examples are not. If we maintain a fixed batch size but reduce the number of unique images in the minibatch, the overall variance in our estimate of the gradient will increase, which we intuitively expect to lead to slower convergence on the training set (we evaluate the gradient variance at initialization for a range of augmentation multiplicities in appendix \ref{app:variance}). To verify this intuition, we plot the cross entropy achieved on the training set after a given number of parameter updates when training the Wide-ResNet on CIFAR-100 at batch size 1024 in Figure \ref{fig:why:a} and batch size 64 in Figure \ref{fig:why:b}. As expected, in both cases smaller augmentation multiplicities achieve faster convergence on the training set. We therefore conclude that the benefits of large augmentation multiplicities arise only when evaluating our model on held-out data, i.e., large augmentation multiplicities impede optimization but benefit generalization.

Indeed, \citet{hoffer2019augment} suggested that large augmentation multiplicities may enable better generalization when the batch size is large because, since gradients evaluated on different augmentations of the same image are correlated, the variance in the final minibatch gradient estimate is larger. They argued that this preserves the generalization benefits of noise observed during small batch training \citep{keskar2016large}. However this argument is incomplete, since it does not give any explanation for why large batch sizes (with augmentation multiplicity) achieve superior test accuracy to smaller batch sizes (without augmentation multiplicity). 

As an alternative explanation, we propose the benefits of augmentation multiplicity arise from the interaction between data augmentation and the use of finite learning rates.\footnote{In support of this claim, we note that in the limit of vanishing learning rates, stochastic gradient descent follows the path of gradient flow for any batch size and any augmentation multiplicity.} Recall that when training with growing batches in Section \ref{sec:growing_batch}, large augmentation multiplicities were stable at larger learning rates, which appeared to account for their superior test accuracy. To investigate whether a similar phenomenon arises when the batch size is fixed, we plot the test accuracy achieved at a range of learning rates when training the Wide-ResNet on CIFAR-100 for 128 epochs at batch size 1024 in Figure \ref{fig:why:c} and at batch size 64 in Figure \ref{fig:why:d}. We observe different behaviour in the small and large batch regimes. For large batch training, large augmentation multiplicities achieve higher test accuracy at similar learning rates, while for small batch sizes large augmentation multiplicities achieve higher test accuracy but require smaller learning rates. In both cases, large augmentation multiplicities do not enable stable training at larger learning rates.

To explain these observations, we note that a large body of work has found that large learning rates have an implicit regularization benefit which enhances the test accuracy, and the strength of this regularization benefit is proportional to the ratio of the learning rate to the batch size \citep{jastrzkebski2017three,li2019towards, smith2021origin}. Reducing the batch size increases the variance in the gradient, which enhances the implicit regularization effect. However prior empirical work has not distinguished whether reducing the batch size enhances the implicit regularization effect by reducing the number of unique examples in the minibatch, or whether it enhances the implicit regularization effect because it reduces the number of samples from the data augmentation process, which increases the variance in the minibatch gradient arising from data augmentation. 

\begin{figure}[t]
\centering
\vskip -3mm
\subfigure[CIFAR-100/large batch]{\includegraphics[width=3.43cm]{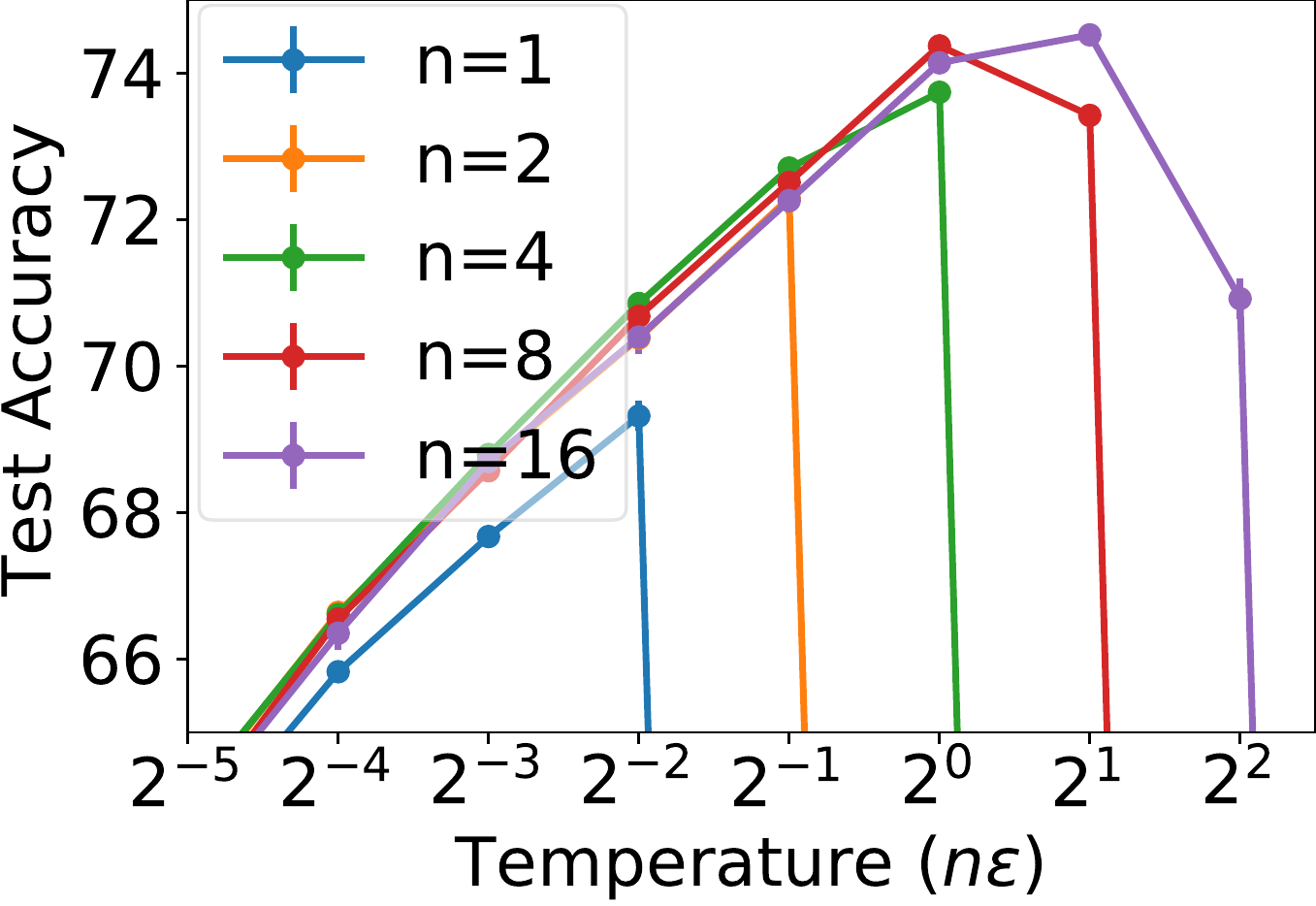}\label{fig:temperature:a}}
\subfigure[CIFAR-100/small batch]{\includegraphics[width=3.43cm]{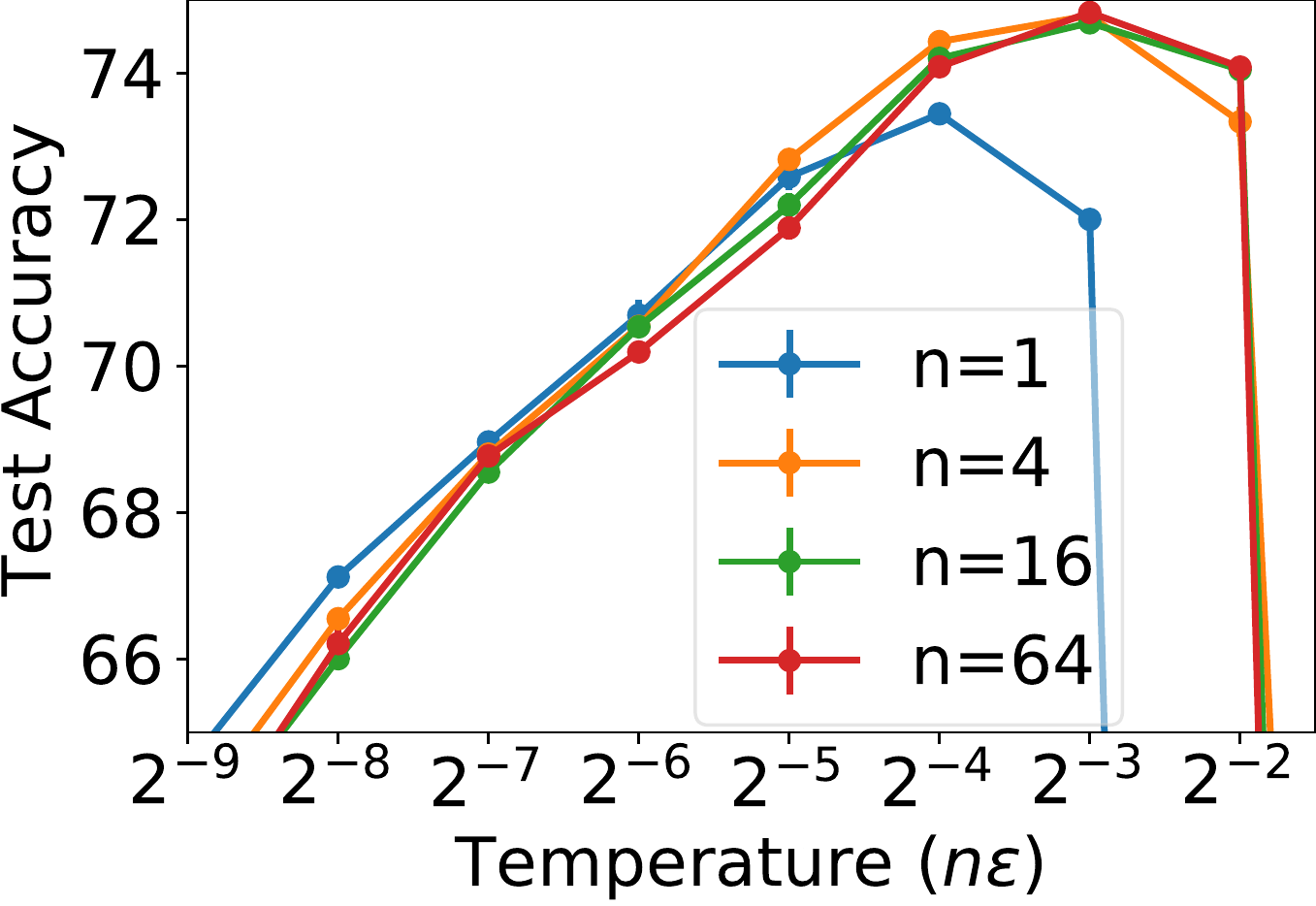}\label{fig:temperature:b}}
\subfigure[ImageNet/large batch]{\includegraphics[width=3.43cm]{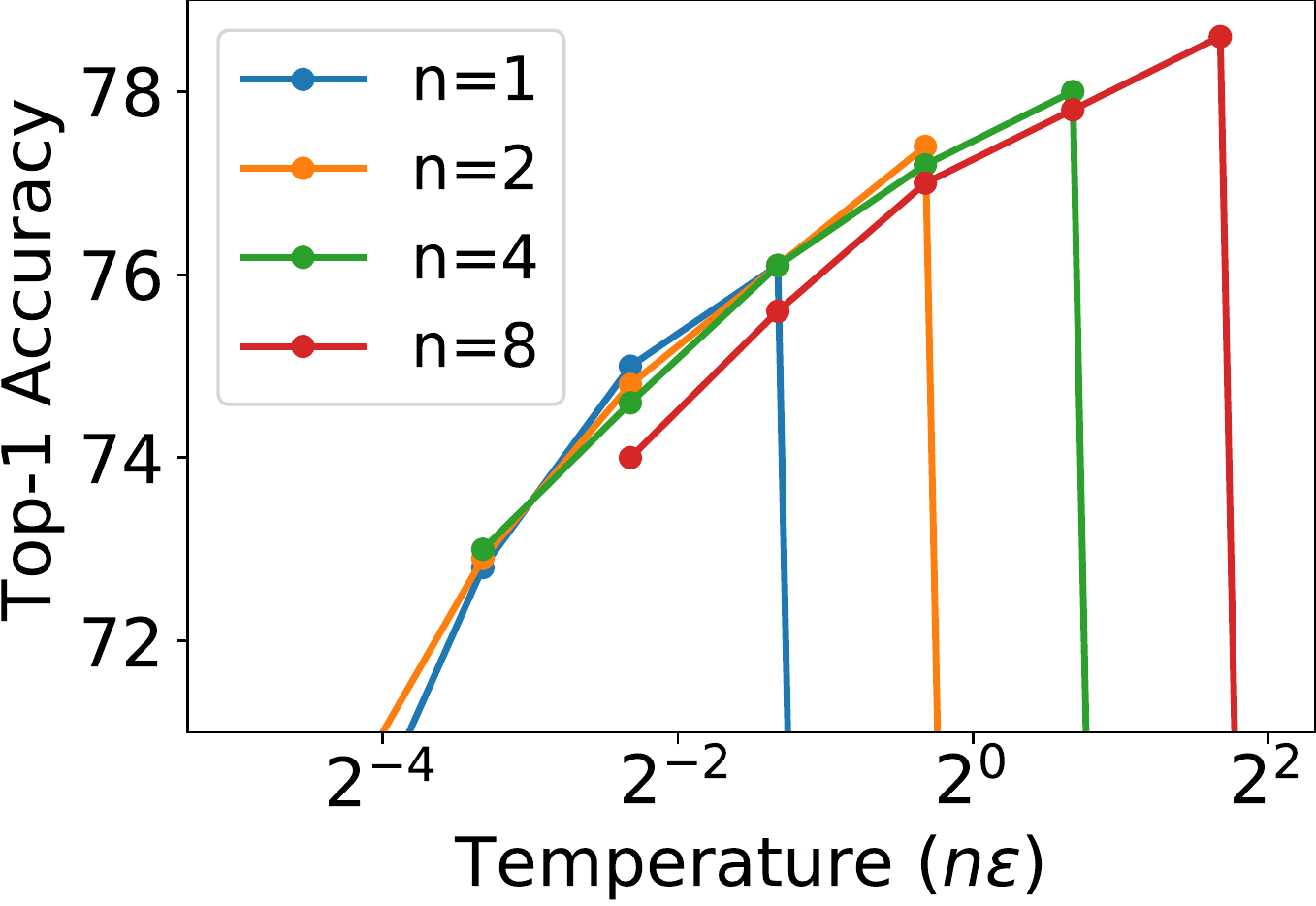}\label{fig:temperature:c}}
\subfigure[ImageNet/small batch]{\includegraphics[width=3.43cm]{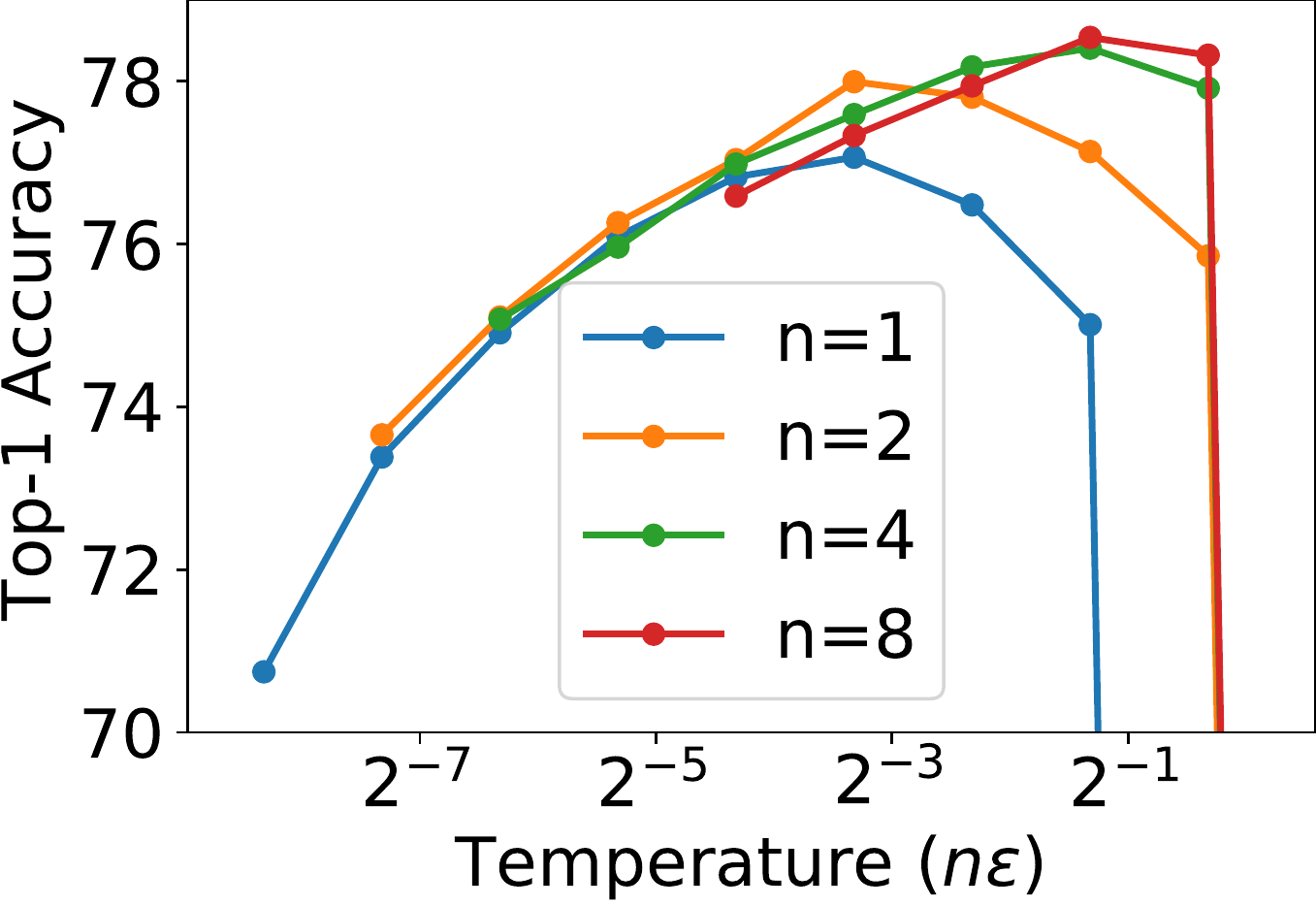}\label{fig:temperature:d}}
 \vskip -2mm
\caption{We plot the test accuracy accuracy at a range of `temperatures' for a 16-4 Wide-ResNet on CIFAR-100 (a/b) and an NF-ResNet50 on ImageNet (c/d) after 128 epochs of training with both large and small batch sizes. We define the temperature as the product of the learning rate $\epsilon$ and the augmentation multiplicity $n$. In all four cases different augmentation multiplicities achieve similar performance when the temperature is small, however larger augmentation multiplicities are stable at higher temperatures, which enables them to achieve higher test accuracies overall.
}
 \vskip -3mm
\label{fig:temperature_main}
\end{figure}

We observed in Section \ref{sec:growing_batch} that reducing the variance arising from the data augmentation procedure increases the test accuracy. We therefore conclude that the implicit regularization benefit of SGD arises from the variance introduced by minibatching, not the variance introduced by data augmentation. On this basis, we predict that when using augmentation multiplicity, the generalization benefit will be governed by the ratio of the learning rate to the number of unique training examples in the batch, not the ratio of the learning rate to the total batch size. When the number of unique examples in each minibatch is fixed (Section \ref{sec:growing_batch}), this ratio is proportional to the learning rate. However, when the batch size is fixed, this ratio is proportional to the product of the learning rate and the augmentation multiplicity, which we define for clarity as the `temperature' \citep{mandt2017stochastic, smith2017bayesian, park2019effect}. In Figure \ref{fig:temperature_main}, we plot the test accuracy achieved after 128 epochs for both the Wide-ResNet/CIFAR-100 and NF-ResNet50/ImageNet in both the small and large batch limits at a range of temperatures. Remarkably, we observe very similar behaviour in each plot. Different augmentation multiplicities achieve similar test accuracy when the temperature is small, but large augmentation multiplicities are stable at larger temperatures, and this enables them to achieve higher test accuracy overall. 

To summarize, large augmentation multiplicities enhance generalization because they suppress a detrimental source of variance in the gradient estimate (the variance arising from data augmentation itself). This enables us to train at larger `temperatures' (either larger learning rates or fewer unique images per minibatch), which in turn enhances the generalization benefit of finite learning rates. This generalization benefit arises from a beneficial source of variance in the gradient estimate (the variance arising from mini-batching). Note that although there is a large body of work studying the implicit regularization benefit of SGD, we believe our paper is the first to distinguish between the different roles of separate sources of variance in the gradient estimate.

%These figures suggest that large augmentation multiplicities achieve higher test accuracy because they suppress the noise in the gradient arising from the data augmentation procedure. This enables stable training at larger temperatures, which enhances the generalization benefit of finite learning rates.

We note however that, as observed in both Section \ref{sec:growing_batch} and Section \ref{sec:fixed}, while large augmentation multiplicities achieve higher test accuracy after tuning the epoch budget, they do achieve lower test accuracy for very large epoch budgets (larger than the optimal epoch budget). Although this regime is not important in practice, this observation suggests that the variance in the gradient arising from data augmentation can help reduce over-fitting when models are over-trained. For completeness, we show in Appendix \ref{app:large_budget_temp} that different augmentation multiplicities do not achieve similar test accuracy at the same temperature for large epoch budgets.

\section{An empirical evaluation of augmentation multiplicity for NFNets}

In Section \ref{sec:fixed}, we showed that large augmentation multiplicities enhance generalization for both small and large batch training. In this section, we verify that these benefits continue to arise for highly performant, strongly regularized models. To achieve this goal, we evaluate the performance of augmentation multiplicity on the NFNet model family, recently proposed by \citet{brock2021high}. NFNets comprise a family of models, denoted by NFNet-Fx, where $x\in \{1,2,3,4,5,6\}$. These models were designed such that it takes roughly twice as long to evaluate a gradient for F2 as for F1 on similar hardware, and similarly twice as long to evaluate a gradient for F3 as for F2, and so on. NFNets are highly expressive, and consequently they benefit from extremely strong regularization and data augmentation, incorporating Dropout \citep{srivastava2014dropout}, Stochastic Depth \citep{huang2016deep}, RandAugment \citep{cubuk2020randaugment}, Mixup \citep{zhang2017mixup} and Cutmix \citep{yun2019cutmix}. They are therefore ideally suited to evaluating the performance of large augmentation multiplicities at scale.

Following the implementation of \citet{brock2021high}, we train each model variant with a batch size of 4096 using SGD with a momentum coefficient of 0.9, cosine annealing \citep{loshchilov2016sgdr}, and Adaptive Gradient Clipping (AGC). A moving average of the weights is stored during training and used to make predictions during inference \citep{szegedy2015going}. We provide results both with and without Sharpness-Aware Minimization (SAM) \citep{foret2020sharpness}, an optimization technique which enhances generalization but roughly doubles the time required to compute a gradient. \citet{brock2021high} use learning rate $\epsilon = 1.6$, however we found $\epsilon = 0.8$ achieves higher top-1 accuracy for large augmentation multiplicities, which we adopt for all multiplicities $n>1$. In the original study, all models were trained for 112,600 parameter updates (corresponding to 360 epochs at augmentation multiplicity 1). For all other details, we refer the reader to \citet{brock2021high}. 

\begin{table}[]
 \vskip -3mm
\caption{NFNet-F2 trained on ImageNet for a range of augmentation multiplicities, without the SAM optimizer, at batch size $4096$. Large augmentation multiplicities achieve higher top-1 accuracy for the same number of parameter updates, and their performance improves further if we increase the compute budget. We also report the performance of NFNet-F3 trained for 225,200 updates at augmentation multiplicity 16. Results marked with an * are taken from the original publication of \citet{brock2021high}.} 
\label{table:f2}
\centering
\resizebox{0.9\textwidth}{!}{
\begin{tabular}{r|c|c|c|c|c|c|c}
\toprule [0.15em]
\multicolumn{1}{c|}{\textbf{}}  & \multicolumn{5}{c|}{\textbf{NFNet-F2 w/out SAM}}                                                                        & \multicolumn{2}{c}{\textbf{NFNet-F3 w/out SAM}}    \\
\midrule [0.1em]
\multicolumn{1}{c|}{\textbf{}} & \multicolumn{5}{c|}{Augmentation Multiplicity:} & 
\multicolumn{2}{c}{Augmentation Multiplicity:} \\
\multicolumn{1}{r|}{Parameter updates:} & 1                          & \multicolumn{1}{c|}{2} & \multicolumn{1}{c|}{4} & \multicolumn{1}{c|}{8} & \multicolumn{1}{c|}{16} &
\multicolumn{1}{c|}{1} &
\multicolumn{1}{c}{16}\\
\midrule [0.1em]
\multicolumn{1}{r|}{112,600}                     & \multicolumn{1}{r|}{85.20} & 85.25                  & 85.35                  & 85.47                  & \textbf{85.49}          & \,\,\,\,\,\,85.7$^*$\,\,\,\,\,\, & \multicolumn{1}{c}{--}           \\
\multicolumn{1}{r|}{225,200}                     & \multicolumn{1}{c|}{--}      &                  \multicolumn{1}{c|}{--}     & 84.53                  & 85.25                  & \textbf{85.71}          & \multicolumn{1}{c|}{--} & \textbf{86.16} \\                
\bottomrule[0.15em]
\end{tabular}
}
\end{table}

\begin{table}[]
\vskip -1mm
\caption{NFNet-F3 and F5, trained on ImageNet at augmentation multiplicity 16 with SAM at batch size 4096. We significantly exceed the original performance reported by \citet{brock2021high}.}
\label{table:sam}
\centering
\resizebox{0.9\textwidth}{!}{
\begin{tabular}{r|c|c|c|c}
\toprule [0.15em]
\multicolumn{1}{c|}{\textbf{Augmentation Multiplicity 16}} & \multicolumn{2}{c|}{\,\,\,\,\textbf{NFNet-F3 w/ SAM}\,\,\,\,} & \multicolumn{2}{c}{\,\,\,\,\textbf{NFNet-F5 w/ SAM}\,\,\,\,} \\
\midrule [0.05em]
Parameter Updates:                                & \,\,112,600\,\,            & 225,200                  & \,\,112,600\,\,            & 168,900                  \\ 
\midrule [0.05em]
\textbf{}                                                  & 85.76              & \textbf{86.45}           & 86.54              & \textbf{86.78} \\          
\bottomrule[0.15em]
\end{tabular}
}
\vskip -2mm
\end{table}

\begin{figure}[t]
\centering
  \vskip -3mm
\subfigure[Tuning NFNet-F2]{\includegraphics[width=0.45\linewidth]{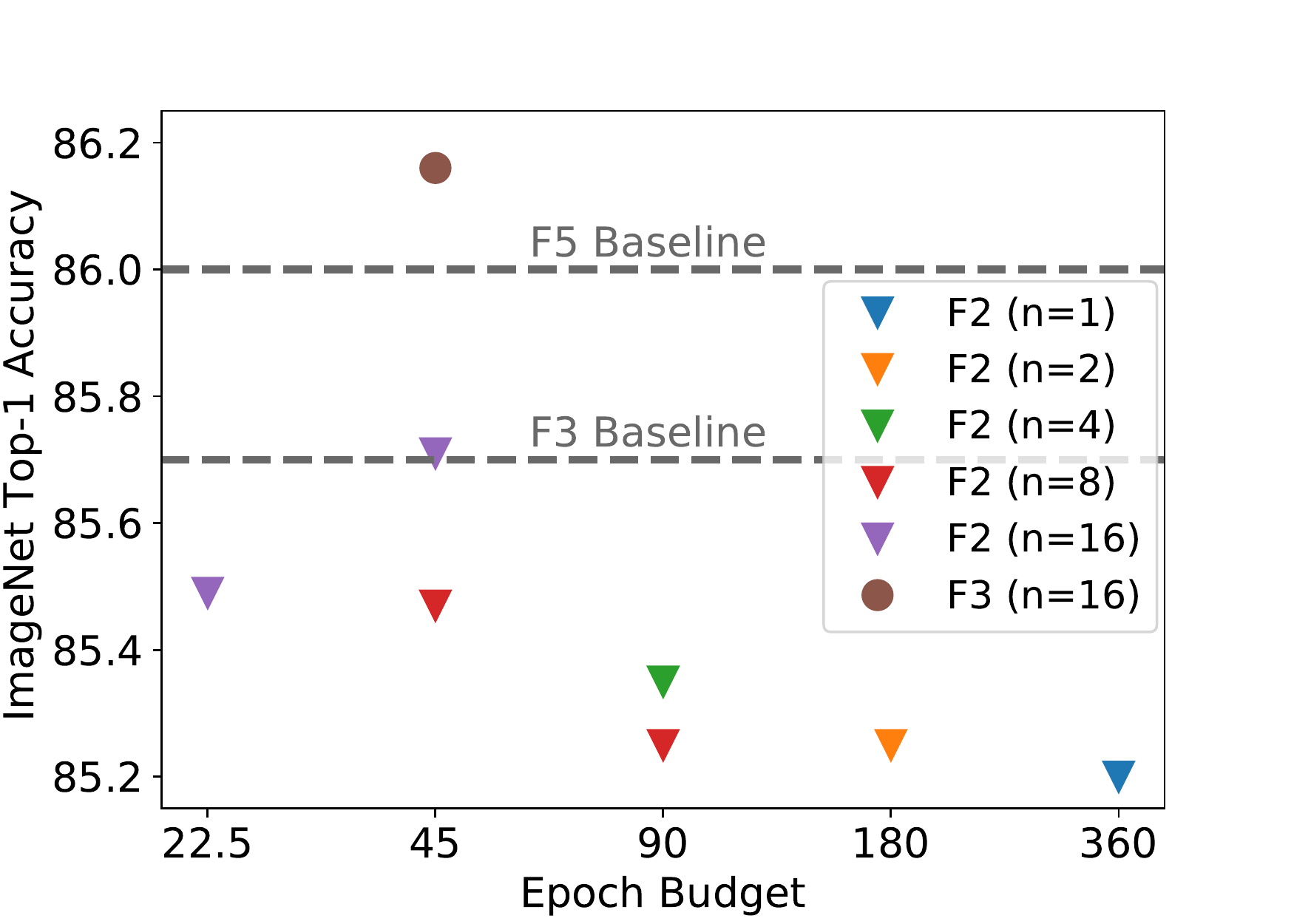}\label{fig:NFNets_F2_diffaugmults_subfigure}} 
% \subfigure[Compute]{\includegraphics[width=0.48\linewidth]{figures/NFNets_compute_id13003032.pdf}\label{fig:NFNets_compute_subfigure}}%
\subfigure[Performance across models]{\includegraphics[width=0.45\linewidth]{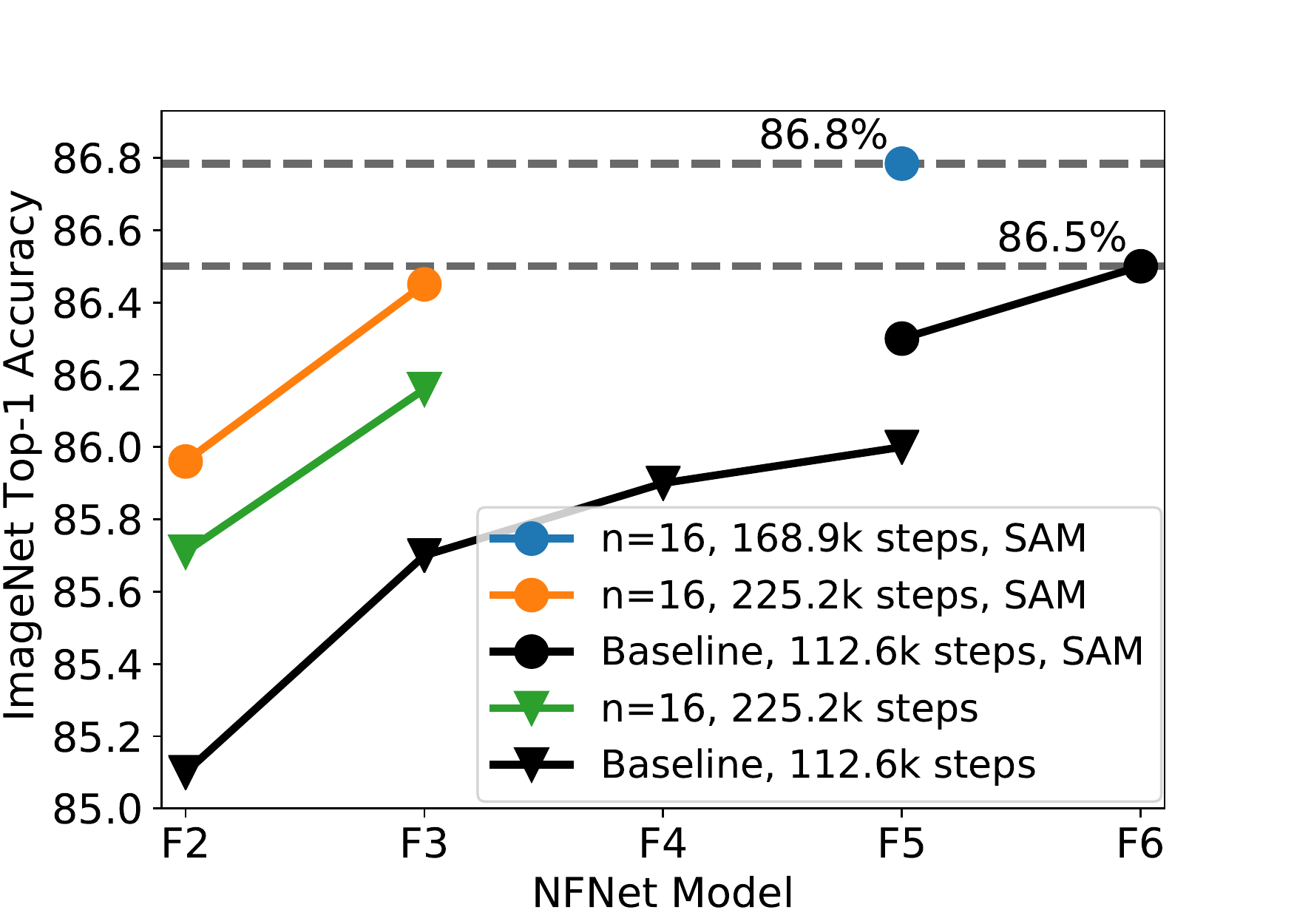}\label{fig:NFNets_params_subfigure}}
\vskip -2mm
\caption{An evaluation of augmentation multiplicity on ImageNet using NFNets. (a) We consider a range of augmentation multiplicities on NFNet-F2 and NFNet-F3, trained at batch size 4096 without the SAM optimizer \citep{foret2020sharpness}. Larger augmentation multiplicities achieve higher top-1 accuracy and require fewer training epochs. For clarity, we emphasize that we define an epoch as a single pass through the training set. (b) We compare the performance achieved at augmentation multiplicity 16 to the original performance reported at augmentation multiplicity 1 by \citet{brock2021high}. Circles denote models trained with SAM, while triangles denote models trained without SAM. Black points denote the original NFNet results reported by \citet{brock2021high} (without augmentation multiplicity). Our models reach substantially higher top-1 accuracy both with and without SAM. }
\label{fig:NFNets_params_and_compute}
\vskip -2mm
\end{figure}

In Table \ref{table:f2}, we provide a detailed evaluation of how augmentation multiplicity influences the performance of NFNet-F2, trained without SAM. Note that when applying augmentation multiplicities larger than 1, we take care to ensure that Mixup and CutMix are applied to different training inputs, not different augmentations of the same image. Larger augmentation multiplicities achieve higher test accuracy, even under a fixed compute budget of 112,600 parameter updates, while augmentation multiplicity 16 continues to improve when the compute budget is increased to 225,200 updates. As we show in Figure \ref{fig:NFNets_F2_diffaugmults_subfigure}, an intriguing implication of these results is that, despite achieving higher test accuracy, large augmentation multiplicities require an order of magnitude fewer training epochs. For clarity, we emphasize that we use `epoch' to denote a full pass through the training set, and that large augmentation multiplicities still require similar numbers of parameter updates. However, this observation demonstrates that we could dramatically reduce the compute cost of training large vision models if we could reduce the variance arising from the data augmentation procedure.

In Table \ref{table:sam} we report the performance of NFNet-F3 and NFNet-F5 when trained with SAM \citep{foret2020sharpness} at augmentation multiplicity 16. Our NFNet-F3 achieves 86.45$\%$ top-1 accuracy after 225,200 updates. This is within $0.05\%$ of the 86.5$\%$ achieved by an NFNet-F6 with SAM after 112,600 updates without augmentation multiplicity in the original paper of \citet{brock2021high}. However since NFNet-F3 requires roughly 8x less time to evaluate a gradient, it achieves this result with roughly 4x less compute, while also containing significantly fewer parameters and being significantly faster to evaluate at inference. After applying augmentation multiplicity, our NFNet-F5 with SAM achieves $86.8\%$ after 168,900 updates. For clarity, we plot a range of results with and without SAM in Figure \ref{fig:NFNets_params_subfigure}. Large augmentation multiplicities consistently outperform the baseline accuracies reported by \citet{brock2021high}.

\section{Discussion}
\label{sec:discussion}

%In our experiments, we study the effect of augmentation multiplicity both by allowing the batch size to grow as the multiplicity increases (keeping the number of unique examples in the minibatch fixed), as well as by reducing the number of unique examples in each minibatch as multiplicity increases (keeping the total batch size fixed). In both cases, we find larger augmentation multiplicities can enhance generalization performance. Remarkably, the higher test set accuracies are achieved in significantly fewer epochs compared to standard training, suggesting it may be possible to significantly speed up training by parallelizing across machines given adequate computational resources.

%Our analysis suggests that the benefits of data augmentation when training deep ResNets \citep{he2016identity} arise because data augmentation introduces a bias in the mini-batch gradients. Although data augmentation also introduces variance in the gradient estimate, after tuning the epoch budget we found that this variance was detrimental to generalization. 

Our study was inspired by a recent study of the origin of the regularization benefits of dropout \citep{wei2020implicit}. Like data augmentation, dropout introduces both bias and variance into minibatch gradients. However \citet{wei2020implicit} found in LSTMs both the bias and the variance introduced by dropout contribute to generalization. For completeness, we investigate the roles of bias and variance in dropout for Wide-ResNets in appendix \ref{app:dropout}. We do not observe a generalization benefit from variance in these experiments.

We have shown that large augmentation multiplicities achieve higher test accuracy, both when the batch size is proportional to the augmentation multiplicity, and when the batch size is fixed (such that the number of unique images in each minibatch decreases as the augmentation multiplicity increases).
A natural question raised by this observation is whether the benefits of augmentation multiplicity arise because we sample multiple augmentations of the same image in the same minibatch, or whether it is sufficient to resample different augmentations of the same unique images in neighbouring minibatches. We briefly compare these two schemes empirically in appendix \ref{app:ablation}, where we find that, while both implementations achieve higher test accuracy than standard training, the benefits of augmentation multiplicity are most significant when we sample multiple augmentations per unique image inside the same minibatch (the scheme studied in the main text). In particular, this approach achieves superior performance for large augmentation multiplicities. 

Finally, we note that augmentation multiplicity has appeared under a range of names in the literature (`batch augmentation' \citep{hoffer2019augment}, `data echoing' \citep{choi2019faster}, `repeated augmentation' \citep{berman2019multigrain}). We chose the terminology `augmentation multiplicity' in this work as it enabled us to describe our experiments more succinctly.

%An inherent limitation of augmentation multiplicity is that it can only be used for networks trained using data augmentation. However the use of data augmentation in deep learning is extremely common, and in particular we note that it is a core component of self-supervised learning \citep{chen2020simple, he2020momentum, grill2020bootstrap}. An additional limitation of our study is that we only consider residual networks on computer vision benchmarks. We leave it to future work to establish whether the benefits of augmentation multiplicity also arise for other model familes and data modalities.

\section{Conclusion}

% \textbf{Conclusion: } 
We provide an empirical study of how augmentation multiplicity, the number of data augmentation samples drawn per unique image in each minibatch, influences test accuracy when training deep residual networks. We find that augmentation multiplicities greater than 1 consistently achieve higher test accuracy, despite achieving slower convergence on the training set (when the overall batch size is fixed). %This benefit arises both when the batch size is proportional to the augmentation multiplicity, and when the batch size is fixed. 
These benefits are particularly significant during large batch training but also arise when the batch size is small. We argue that this phenomenon arises from the interaction between the variance data augmentation introduces into the gradient estimate and the role of finite learning rates during training.
Our study suggests practitioners should consider choosing large augmentation multiplicities as the default when training neural networks for computer vision.

 \section*{Acknowledgements}
We thank Yee Whye Teh, Karen Simonyan, Zahra Ahmed and Hyunjik Kim for helpful advice, and Matthias Bauer for feedback on an earlier draft of the manuscript.

\newpage

\bibliography{augmult.bib}
\bibliographystyle{plainnat}

%\clearpage
%\onecolumn 

\appendix

\section{Augmentation multiplicity and gradient variance}
\label{app:variance}

\begin{figure}[h]
\centering
\includegraphics[height=4.0cm]{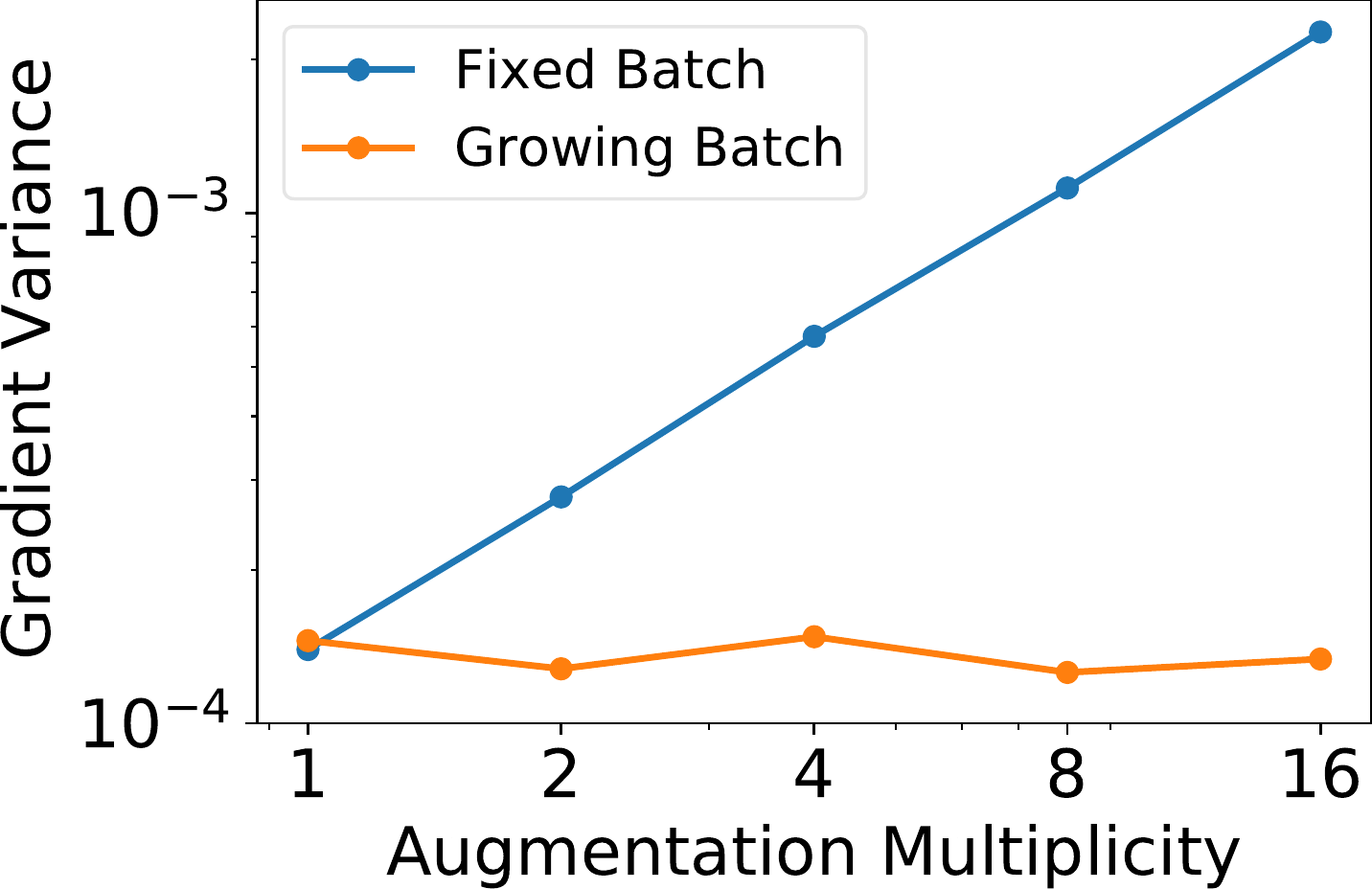}
% \vskip -3mm
\caption{We evaluate the mean variance in the minibatch gradient estimate, across all model parameters for a 16-4 Wide-ResNet evaluated on CIFAR-100 at initialization without weight decay. For ``fixed batch'' the batch size $B=16$ is independent of the augmentation multiplicity, while for ``growing batch'' the batch size $B=16n$ grows as the augmentation multiplicity rises to ensure the number of unique examples in each minibatch is fixed. In the ``growing batch'' scheme, the variance across minibatches does not depend strongly on the augmentation multiplicity, however in the ``fixed batch'' scheme, the variance across minibatches increases as the augmentation multiplicity increases.
}
% \vskip -2mm
\label{fig:variance}
\end{figure}

To verify that increasing the augmentation multiplicity increases the variance in the minibatch estimate of the gradient when the batch size is fixed, we plot the variance between minibatch gradients for the 16-4 Wide-ResNet at initialization in figure \ref{fig:variance}. Since the gradient at initialization is otherwise dominated by the $L_2$ loss, we set the $L_2$ coefficient to zero. The model is otherwise unchanged. We first evaluate the variance for all model parameters, then average the variance across all parameters within a particular layer, before averaging again across all model layers to obtain a single scalar value. When the number of unique examples in each minibatch is fixed (``growing batch''), the variance between minibatches does not depend strongly on the augmentation multiplicity, however when the batch size is fixed (``fixed batch''), the variance increases as the augmentation multiplicity rises.

\section{The dependence on the temperature for large epoch budgets}
\label{app:large_budget_temp}

\begin{figure}[h]
\centering
\subfigure[Large batch/32 epochs]{\includegraphics[height=2.36cm]{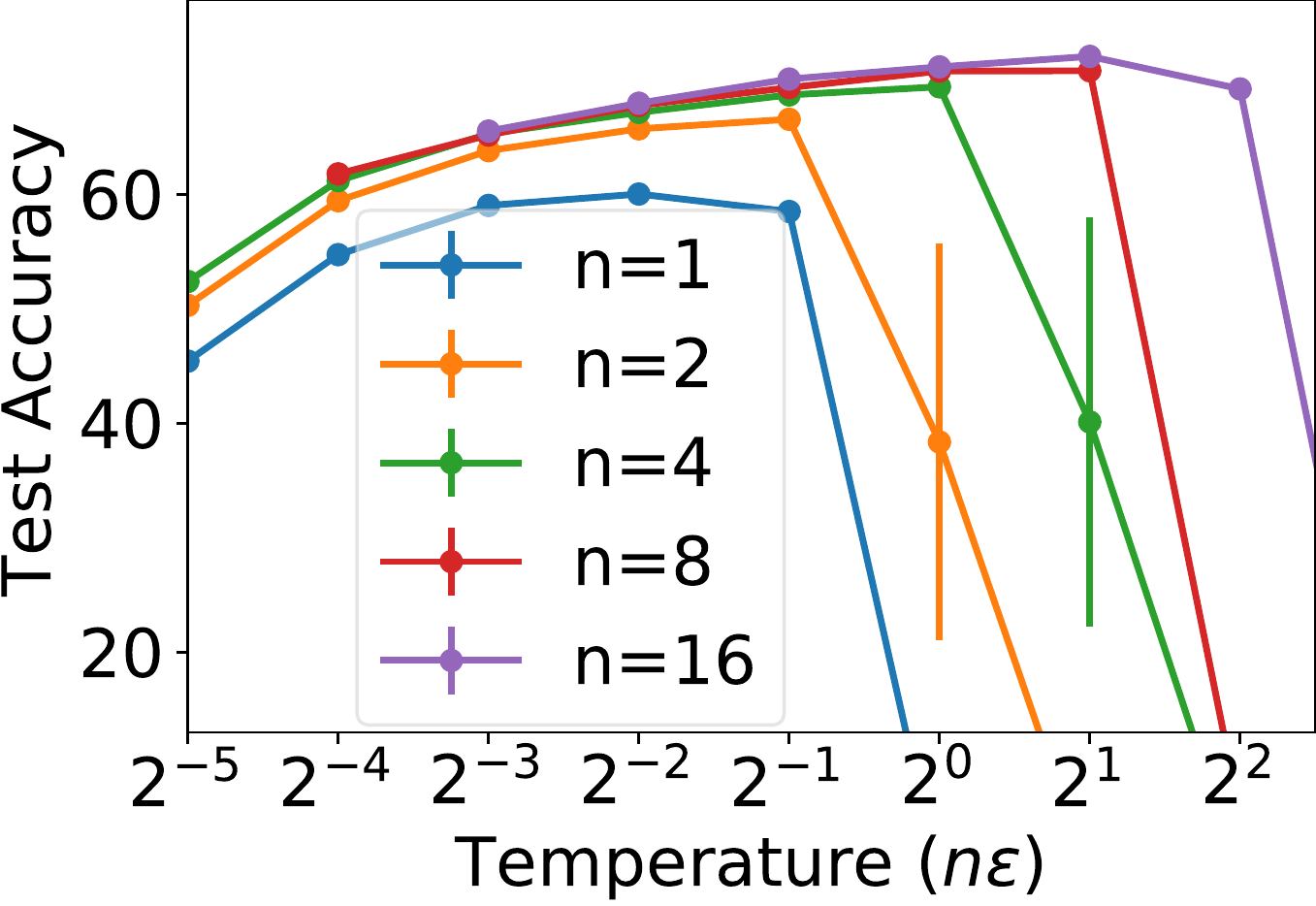}\label{fig:large_32}}
\subfigure[Small batch/32 epochs]{\includegraphics[height=2.36cm]{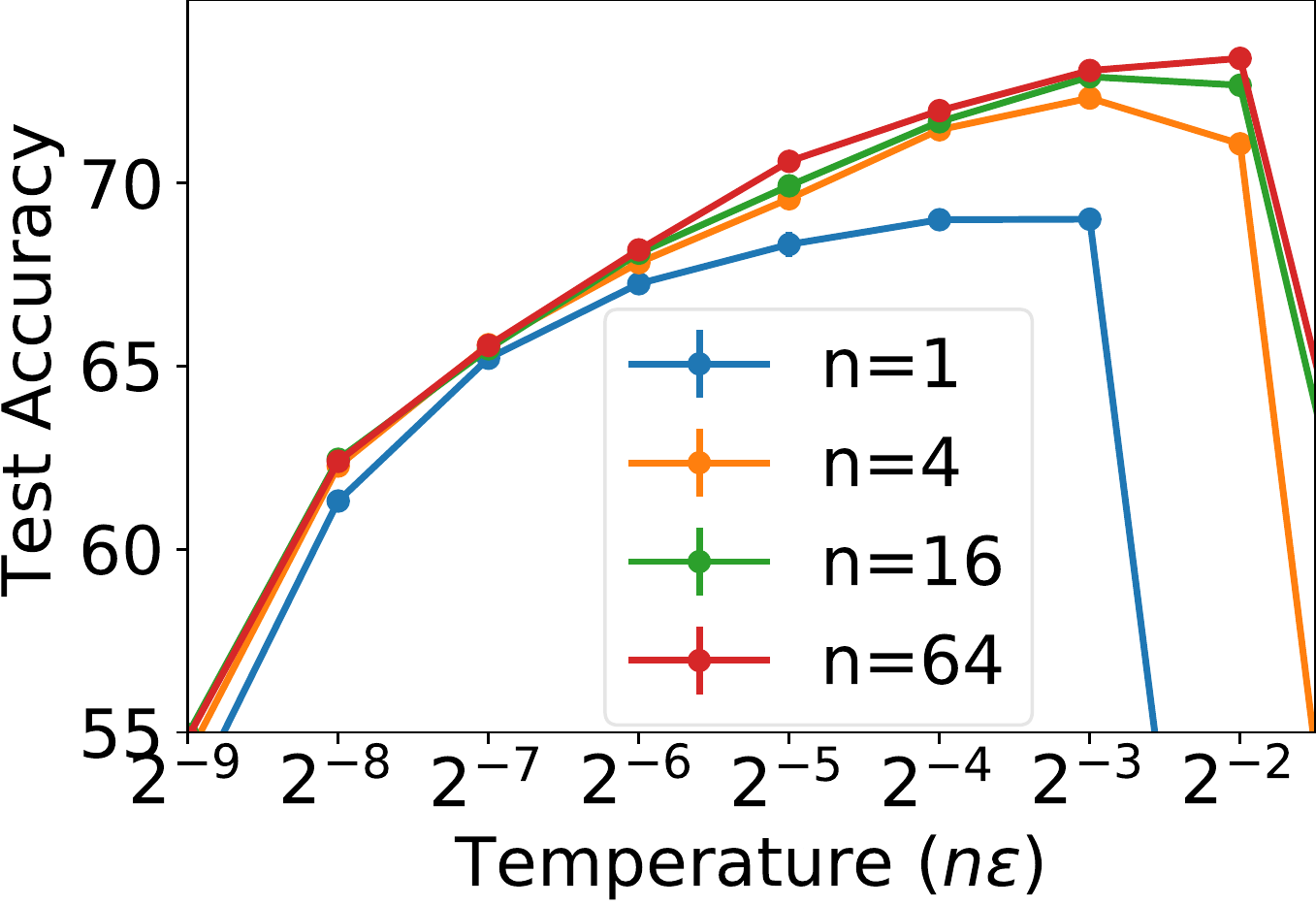}\label{fig:small_32}}
\subfigure[Large batch/512 epochs]{\includegraphics[height=2.36cm]{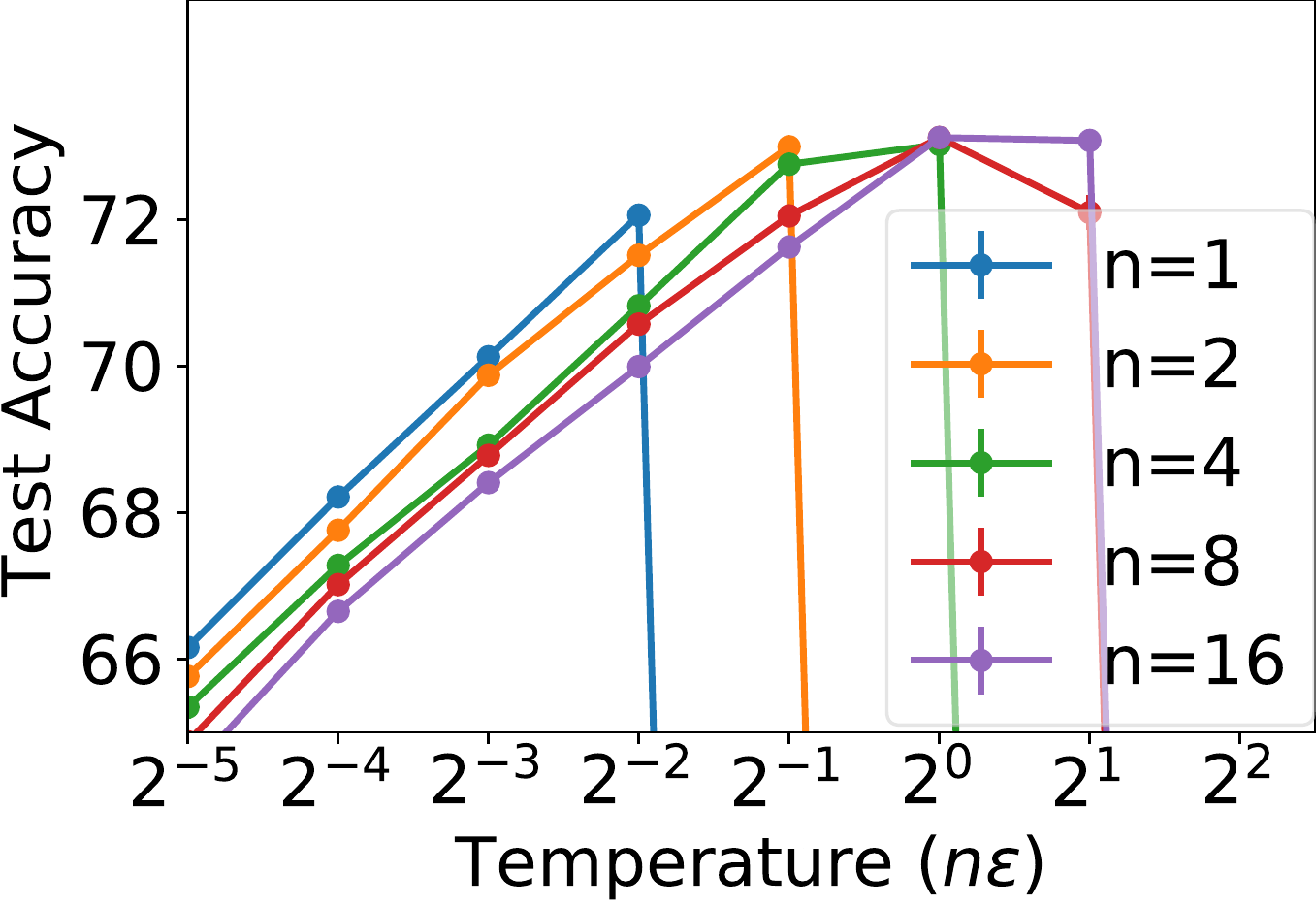}\label{fig:large_512}}
\subfigure[Small batch/512 epochs]{\includegraphics[height=2.36cm]{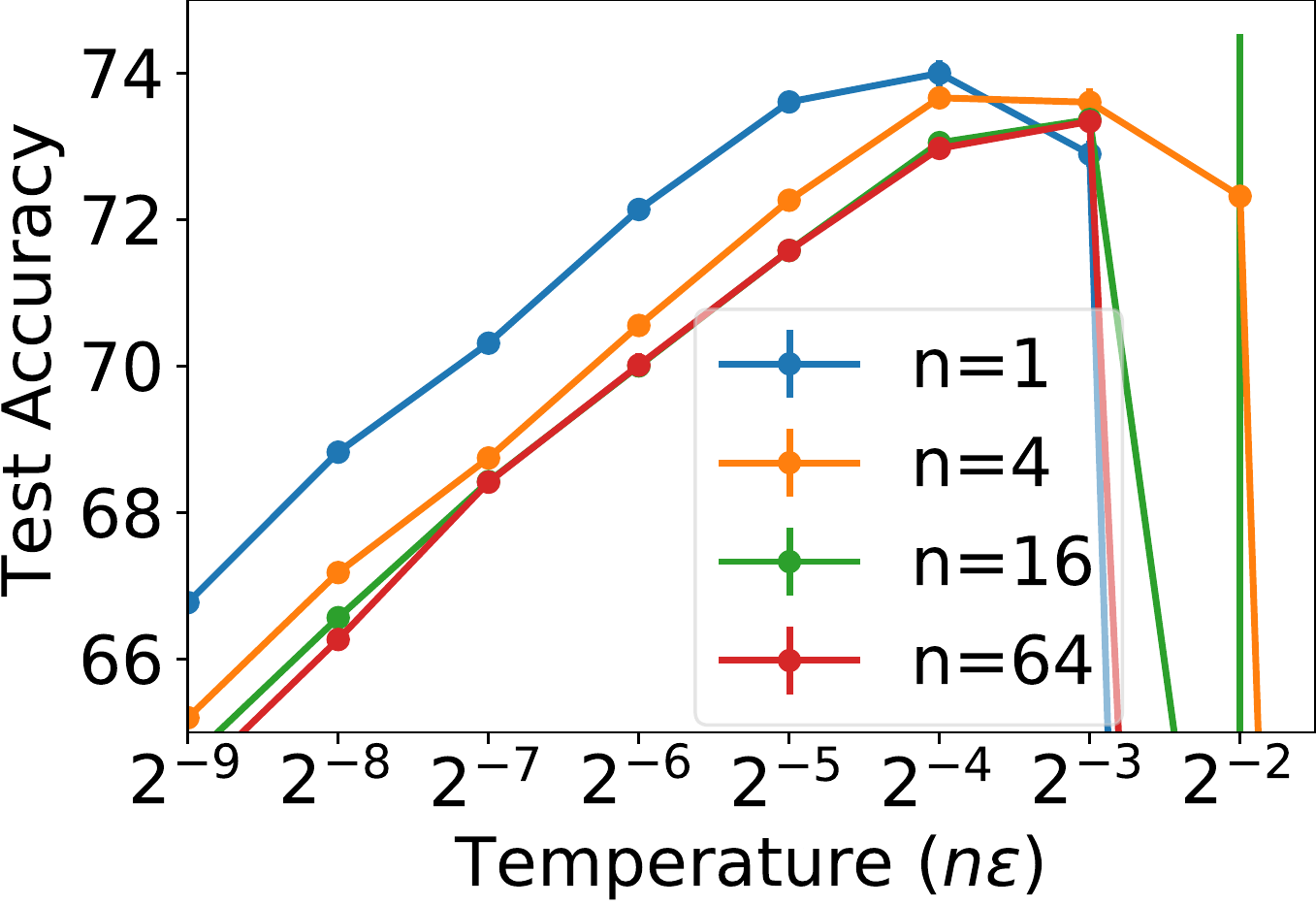}\label{fig:small_512}}

% \vskip -3mm
\caption{A 16-4 Wide-ResNet trained on CIFAR-100 at two batch sizes, 1024 (large) and 64 (small). In (a)/(b) we exhibit how the test accuracy depends on the temperature, defined as the product of the augmentation multiplicity $n$ and the learning rate $\epsilon$, for a budget of 32 training epochs. For both batch sizes, we find that the test accuracy is determined by the temperature for small epoch budgets when the temperature is also small, while large augmentation multiplicities achieve higher test accuracy for large temperatures. Meanwhile in (c)/(d), we show that small augmentation multiplicities achieve higher test accuracy at a given temperature when the epoch budget is large (512 training epochs).
}
% \vskip -2mm
\label{fig:large_epoch}
\end{figure}

We showed in Section \ref{sec:analysis} that when the batch size is fixed, different augmentation multiplicities achieve similar test accuracy for moderate epoch budgets at small `temperatures', which we define as the product of the augmentation multiplicity $n$ and the learning rate $\epsilon$, while large augmentation multiplicities achieve higher test accuracy for large temperatures. However we also observed that for large epoch budgets (when models are over-trained), small augmentation multiplicities can reduce overfitting. To explore this further, in Figure \ref{fig:large_epoch} we show how the test accuracy depends on the temperature for both small and large epoch budgets on our 16-4 Wide-ResNet/CIFAR-100. We find that at small epoch budgets, different augmentation multiplicities achieve similar test accuracy when the temperature is small, as observed for moderate epoch budgets in Section \ref{sec:analysis}, while for large epoch budgets, small augmentation multiplicities achieve higher test accuracy at a given temperature, verifying that small augmentation multiplicities can help prevent overfitting in over-trained models.

\section{Do multiple augmentations need to occur within the same minibatch?}
\label{app:ablation}
\begin{figure}[t]
\centering
  \vskip -1mm
\subfigure[]{\includegraphics[height=4.6cm]{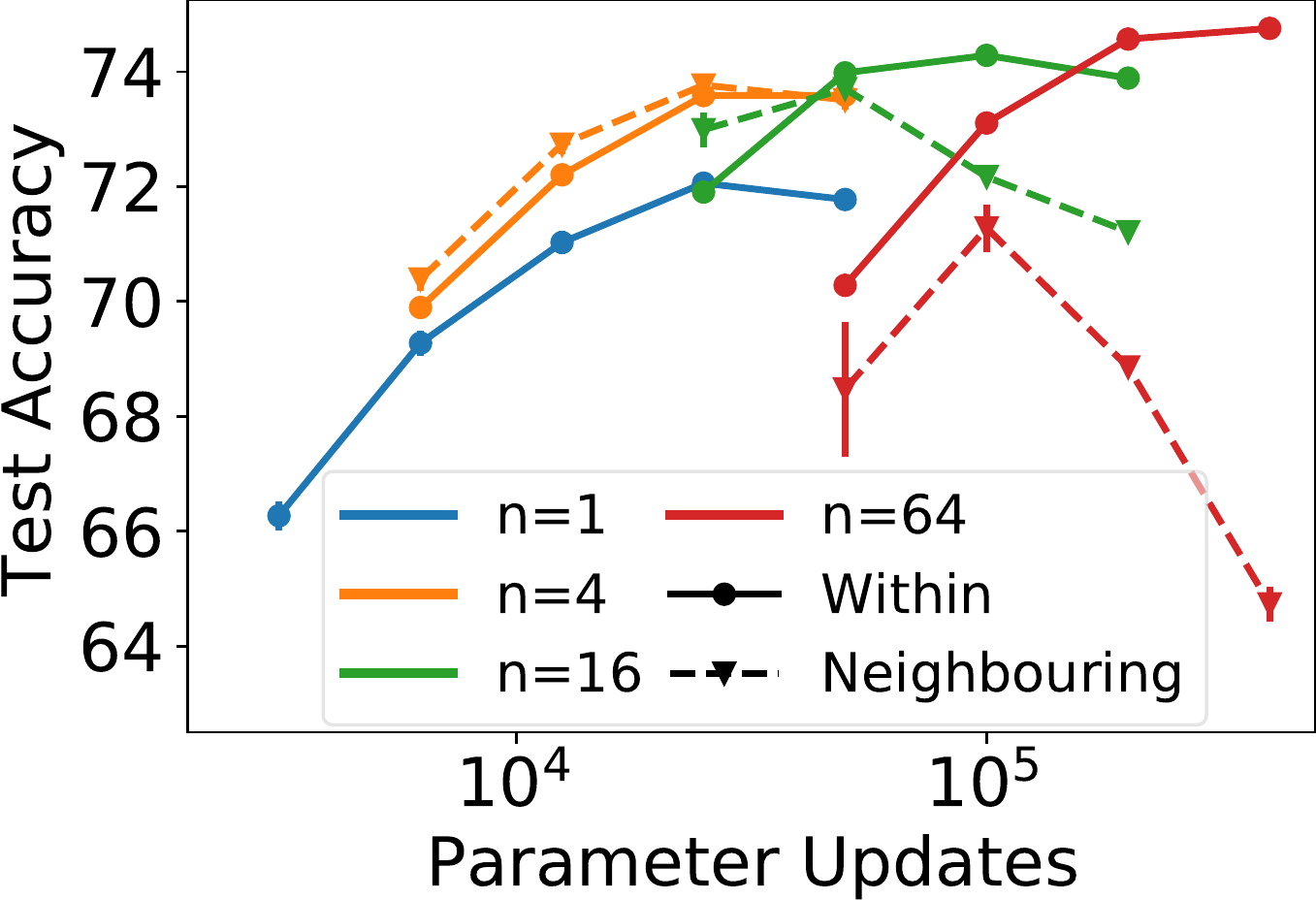}\label{fig:ablation_a}}
\subfigure[]{\includegraphics[height=4.6cm]{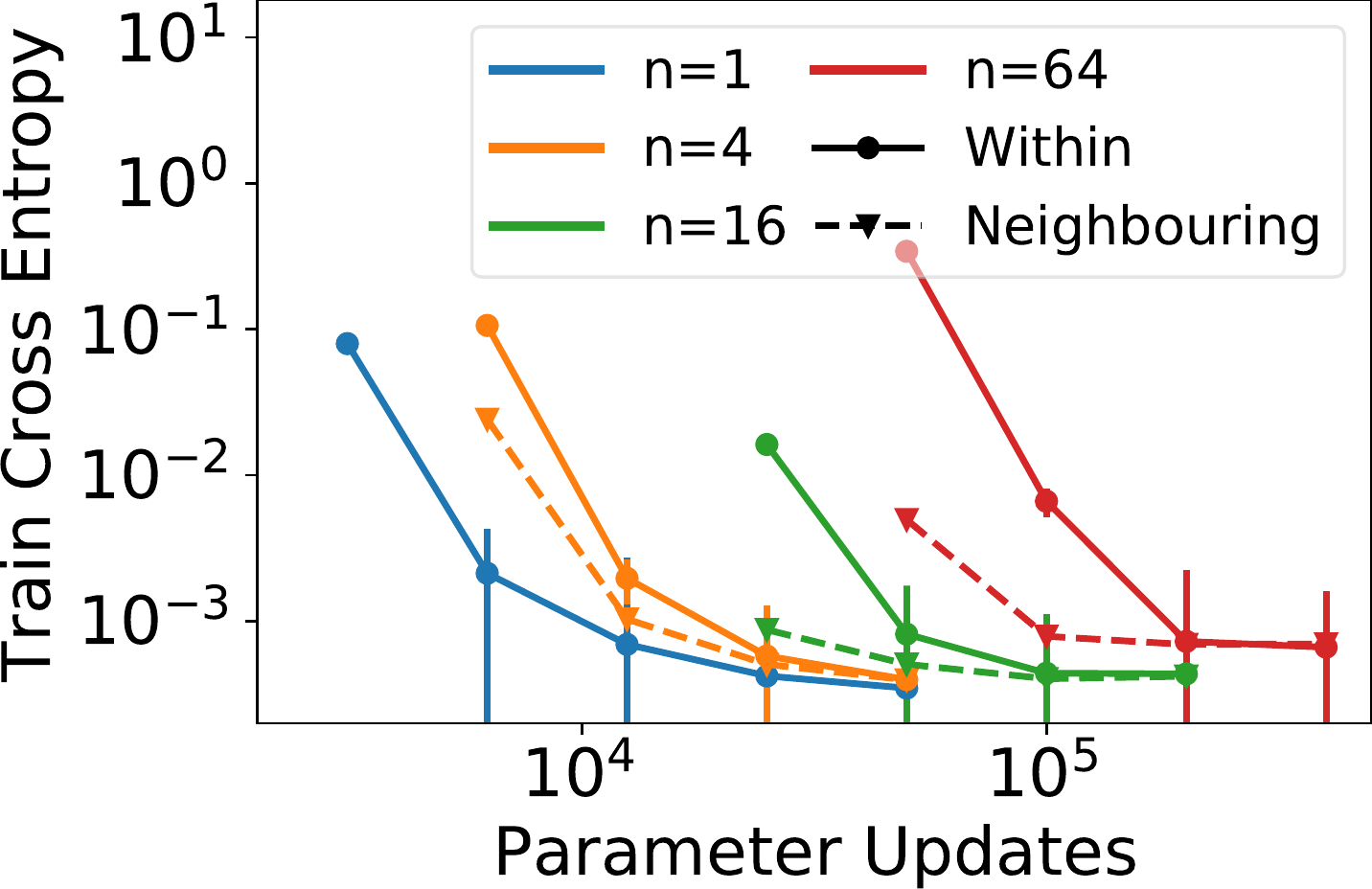}\label{fig:ablation_b}}

  \vskip -2mm
\caption{A 16-4 Wide-ResNet trained on CIFAR-100 with a fixed batch size of 1024. Solid lines (``within'') denote the default implementation of augmentation multiplicity, whereby we sample $n$ augmentations per image within each minibatch (reducing the number of unique images per batch). Meanwhile dashed lines (``neighbouring'') denote an alternative scheme, whereby we only sample a single augmentation per image in each minibatch, but we resample the same training examples in $n$ neighbouring minibatches (with different data augmentations). (a) We find that the neighbouring scheme can also enhance generalization for moderate multiplicities, however it performs poorly when the multiplicity is large. (b) The ``neighbouring'' scheme also achieves slightly lower training losses.}
\label{fig:ablation}
  \vskip -2mm
\end{figure}

In the main text, we analysed a single version of augmentation multiplicity. In this default implementation, we draw $n$ augmentations of each unique image \textit{within the same minibatch}. When the batch size is fixed, this implies the number of unique images per batch decreases as the augmentation multiplicity increases. In Figure \ref{fig:ablation} we compare this default implementation (which we refer to as ``within'') to an alternative scheme, whereby we only sample a single augmentation per unique image in any single minibatch, but we sample the same set of unique images in $n$ neighbouring minibatches. For clarity, these minibatches contain the same unique images but we draw different augmentations of each image in each adjacent minibatch. We refer to this alternative scheme as ``neighbouring''.

Normal training (augmentation multiplicity 1) is shown in blue, the default implementation of augmentation multiplicity (``within'') is shown with full lines, and the alternative implementation (``neighbouring'') is shown with dashed lines. For moderate multiplicities, both implementations enhance the test accuracy, as shown in Figure \ref{fig:ablation_a}. However for large augmentation multiplicities, the default implementation (``within'') continues to enhance generalization, while the performance of the alternative scheme (``neighbouring'') begins to degrade. We therefore conclude that to maximize the benefits of augmentation multiplicity, one should adopt the default implementation (``within''), placing multiple augmentations of the same image inside the same minibatch.

\section{Bias and variance in dropout}
\label{app:dropout}
Dropout \citep{srivastava2014dropout} is a widely used regularization technique in deep learning. For each minibatch gradient evaluation, a random boolean mask is sampled which covers the output activations of a chosen layer of the network, such that any output unit (and its connections) is dropped with a specified drop probability. The purpose of dropout is to force the network to learn redundant representations, thus discouraging overfitting and enhancing generalization.

Like data augmentation, dropout has two influences on the gradient distribution. First, it changes the expected value of the gradient, introducing bias. Second, since we typically only sample a single dropout mask, dropout also introduces a source of variance. The roles of these two effects was studied by \citet{wei2020implicit}, who refer to the bias and variance as the implicit and explicit regularization effects respectively. To disentangle the role of bias and variance, they take an average of $n$ gradients evaluated for $n$ different random dropout masks before taking each optimization step. The larger the $n$, the lower the variance arising from the stochasticity of sampling a dropout mask, while the bias remains unchanged. The authors found that both the bias and the variance arising from dropout enhance generalization in LSTMs. Note that by contrast, we found that the variance arising from the data augmentation procedure harms generalization for ResNets trained on CIFAR-100 and ImageNet.

\begin{figure}[t]
\centering
  \vskip -1mm
\subfigure[]{\includegraphics[height=3.0cm]{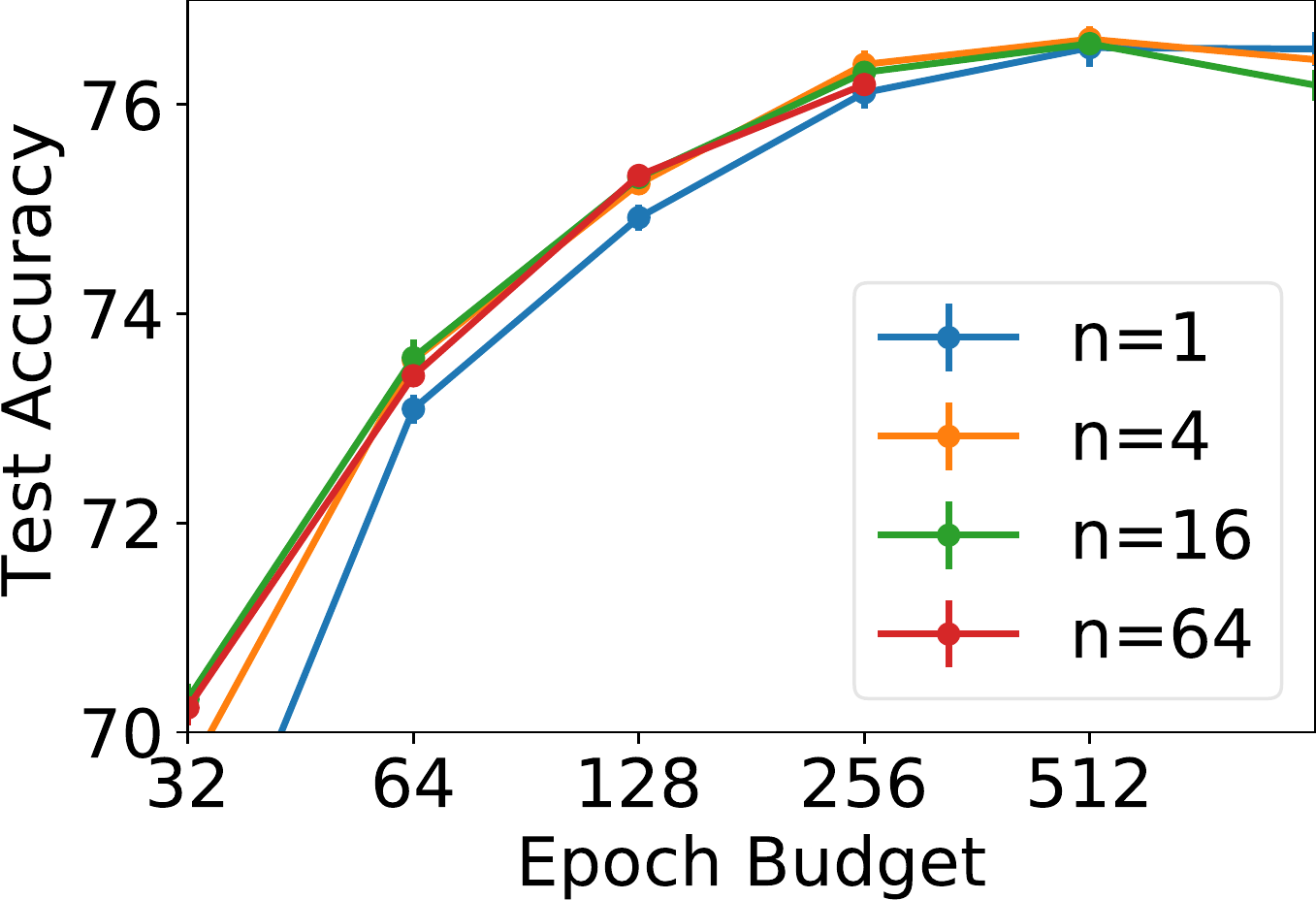}\label{fig:dropout_a}}
\subfigure[]{\includegraphics[height=3.0cm]{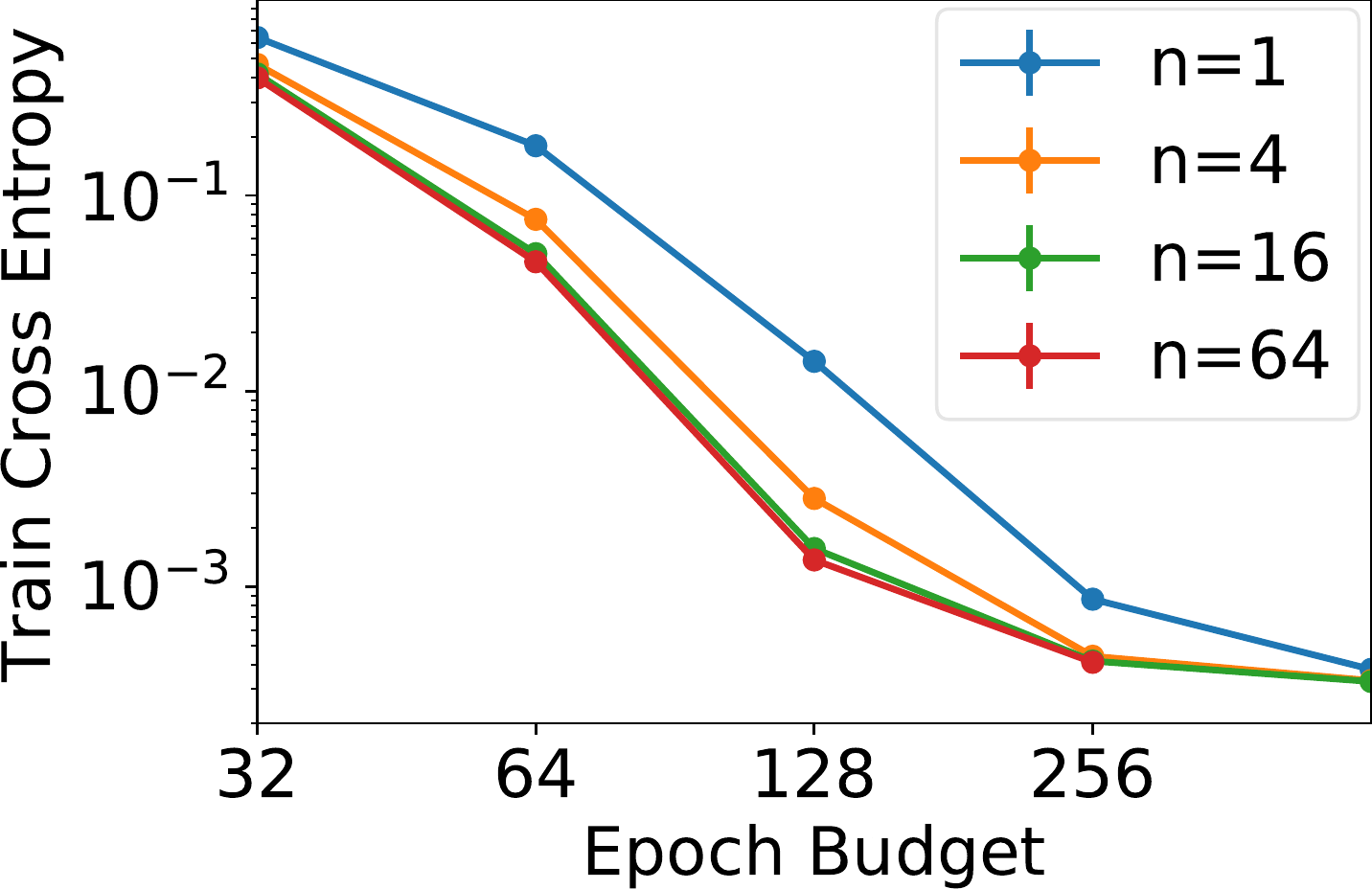}\label{fig:dropout_b}}
\subfigure[]{\includegraphics[height=3.0cm]{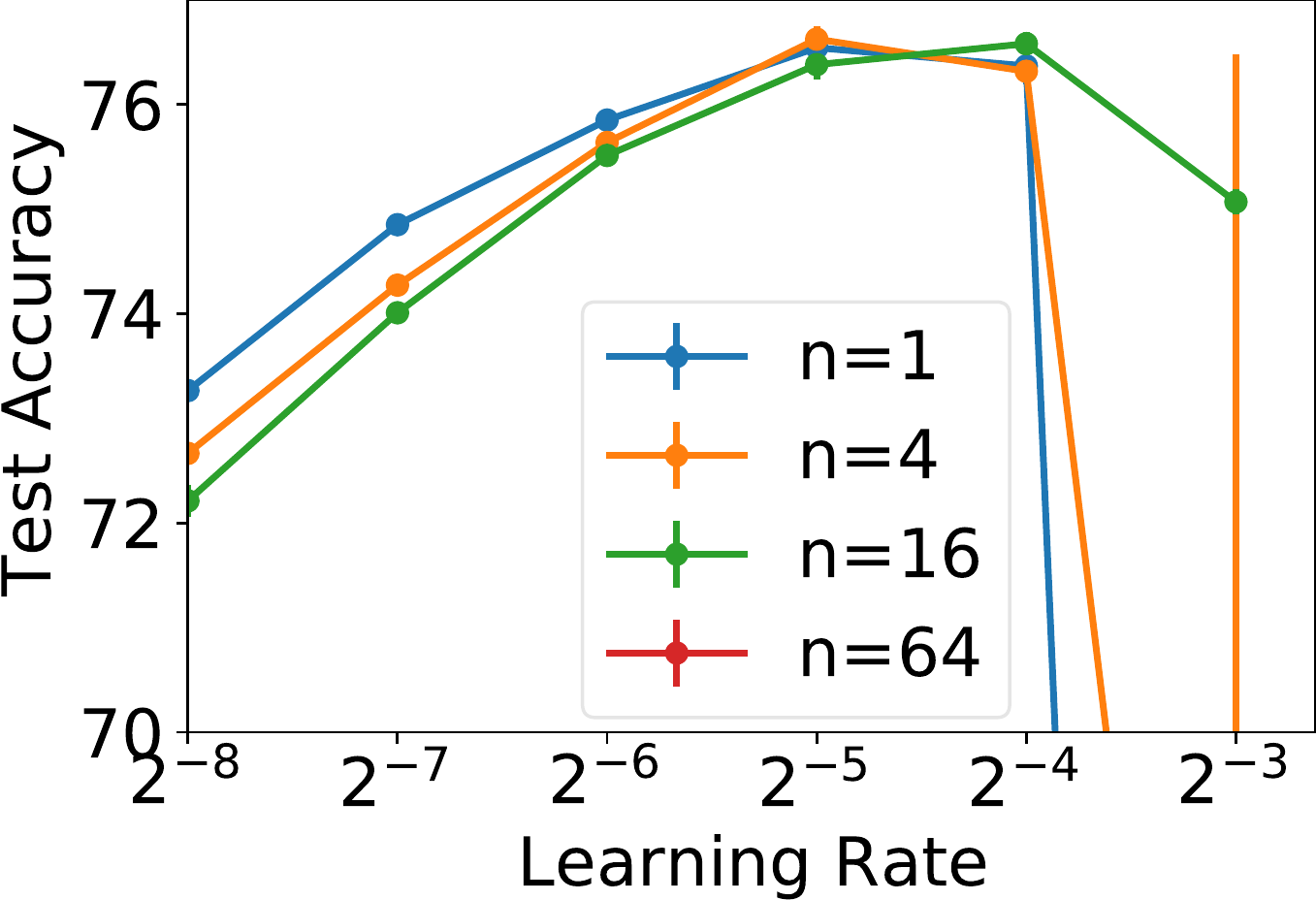}\label{fig:dropout_c}}
  \vskip -2mm
\caption{Results with a 16-4 Wide-ResNet on CIFAR-100, trained using dropout with a drop probability of $0.4$. The number of unique images in each batch is fixed at 64, and we evaluate the performance at augmentation multiplicity 1. However we average each gradient across $n$ models generated from $n$ different samples of the dropout mask. In figure (a), we find that the test accuracy does not depend strongly on the number of samples of the dropout mask, which suggests that the benefits of dropout in this network arise from the bias introduced in the gradient, not the variance. In figure (b) we show that large numbers of samples minimize the training loss more quickly. Finally in figure (c) we provide the test accuracy achieved for a range of learning rates after 512 epochs.}
\label{fig:dropout}
  \vskip -2mm
\end{figure}

We replicate the empirical analysis of \citet{wei2020implicit} for a 16-4 Wide-ResNet trained on CIFAR-100 in Figure \ref{fig:dropout}. We apply dropout on the final linear layer of the network with drop probability $0.4$, averaging each gradient over $n$ samples of the dropout mask. Note that we average the gradient across different dropout masks but with the same augmentations of the input (i.e., the augmentation multiplicity is set to 1 throughout). We find that networks trained with dropout achieve higher test accuracy, while also requiring larger epoch budgets (see Figure \ref{fig:small_a} for an equivalent network trained without dropout at a range of augmentation multiplicities). However reducing the dropout variance by averaging the gradient across multiple masks does not appear to substantially enhance test accuracy, suggesting that the benefits of dropout in this network primarily arise from the bias in the gradient.

%\section{NFNet Augmentation Details} 

\end{document}